\documentclass[letterpaper]{article} 
\usepackage[preprint]{aaai2027} 
\usepackage[hyphens]{url}  
\usepackage{graphicx} 
\usepackage{natbib}  
\usepackage{caption} 
\usepackage{algorithm}
\usepackage{algorithmic}
\usepackage{booktabs}
\usepackage{amsmath}
\usepackage{amssymb}
\usepackage[table]{xcolor}
\usepackage{array}
\usepackage{makecell}

\usepackage{tabularx}
\usepackage{array}

\newcommand{\logtext}[1]{%
{\raggedright\ttfamily\scriptsize #1\par}%
}

\definecolor{asrblue}{RGB}{72, 110, 150}
\definecolor{asrrow}{RGB}{244, 248, 252}
\definecolor{backbonerow}{RGB}{232, 240, 248}
\definecolor{realblue}{RGB}{96, 126, 142}
\definecolor{realrow}{RGB}{235, 244, 245}
\definecolor{realbackbone}{RGB}{218, 235, 237}

\title{A-SR: Self-Evolving Agentic LLMs for Symbolic Regression via Hierarchical Coordination}

\author{
    Wenxiao Zhao\textsuperscript{\rm 1,2},
    Dong Liu\textsuperscript{\rm 2},
    Kaiyi Xu\textsuperscript{\rm 1},
    Feng Liu\textsuperscript{\rm 1},
    Zhen Zhao\textsuperscript{\rm 1},
    Fei Ben\textsuperscript{\rm 1},\\
    Shu Wang\textsuperscript{\rm 2},
    Wenhao Li\textsuperscript{\rm 3},
    Ying Nian Wu\textsuperscript{\rm 2},
    Fenghua Ling\textsuperscript{\rm 1},
    Haobo Li\textsuperscript{\rm 1},
    Lei Bai\textsuperscript{\rm 1}
}
\affiliations{
    \textsuperscript{\rm 1}Shanghai AI Laboratory\\
    \textsuperscript{\rm 2}University of California at Los Angeles\\
    \textsuperscript{\rm 3}Tongji University\\
}

\begin{document}

\maketitle

\begin{abstract}
Symbolic regression aims to discover closed-form equations from data, yet existing LLM-guided methods often rely on a unified proposal loop that compresses heterogeneous search failures into a scalar score and a single prompt. We propose A-SR, a self-evolving agentic framework that shifts the control unit from expression edits to role-conditioned evidence views. A-SR coordinates formula discovery through routing among coordination protocols, online evaluator-reward role policy, and state-routed process memory. During search, evaluator feedback is used to characterize reliability and productivity, update role-level utilities, and route elite motifs, failure traces, and validity diagnostics to different agents. The framework self-evolves at two timescales: within a run, it adapts the search process without updating LLM parameters; across runs, recorded trajectories can be distilled into open-source LLMs as role-conditioned proposal priors. Averaged over the four LSR-Synth scientific domains in LLM-SRBench, A-SR improves Acc@0.01 over baselines from 25.79\% to 48.30\% with Llama3.1-8B, while A-SR-LoRA improves the corresponding Qwen3-4B result from 24.58\% to 38.29\%. On four real-world scientific discovery tasks, A-SR obtains the best ID/OOD NMSE on 7 of 8 reported metrics.
\end{abstract}

\section{Introduction}

Scientific discovery often seeks compact mathematical laws that are accurate on observed data, executable as symbolic expressions, and robust under extrapolation \citep{bongard2007automated,schmidt2009distilling,udrescu2020ai,cranmer2020discovering}. Symbolic regression (SR) addresses this goal by searching for closed-form expressions. Prior SR systems have approached this search through genetic programming, physics-inspired equation discovery, neural-guided expression generation, and hybrid evolutionary-neural search, including Eureqa, AI Feynman, PySR, DSR/uDSR, and transformer-based SR models \citep{schmidt2009distilling,udrescu2020ai,cranmer2023pysr,petersen2021deep,landajuela2022unified,biggio2021neural,kamienny2022end}. However, practical SR systems must handle invalid programs, unstable parameterizations, redundant terms, and poor out-of-domain behavior. These failures are not merely different degrees of one objective: they require different evidence and corrective actions, yet many systems compress them into one scalar reward and one proposal prompt.

\begin{figure}[t]
\centering
\includegraphics[width=\columnwidth]{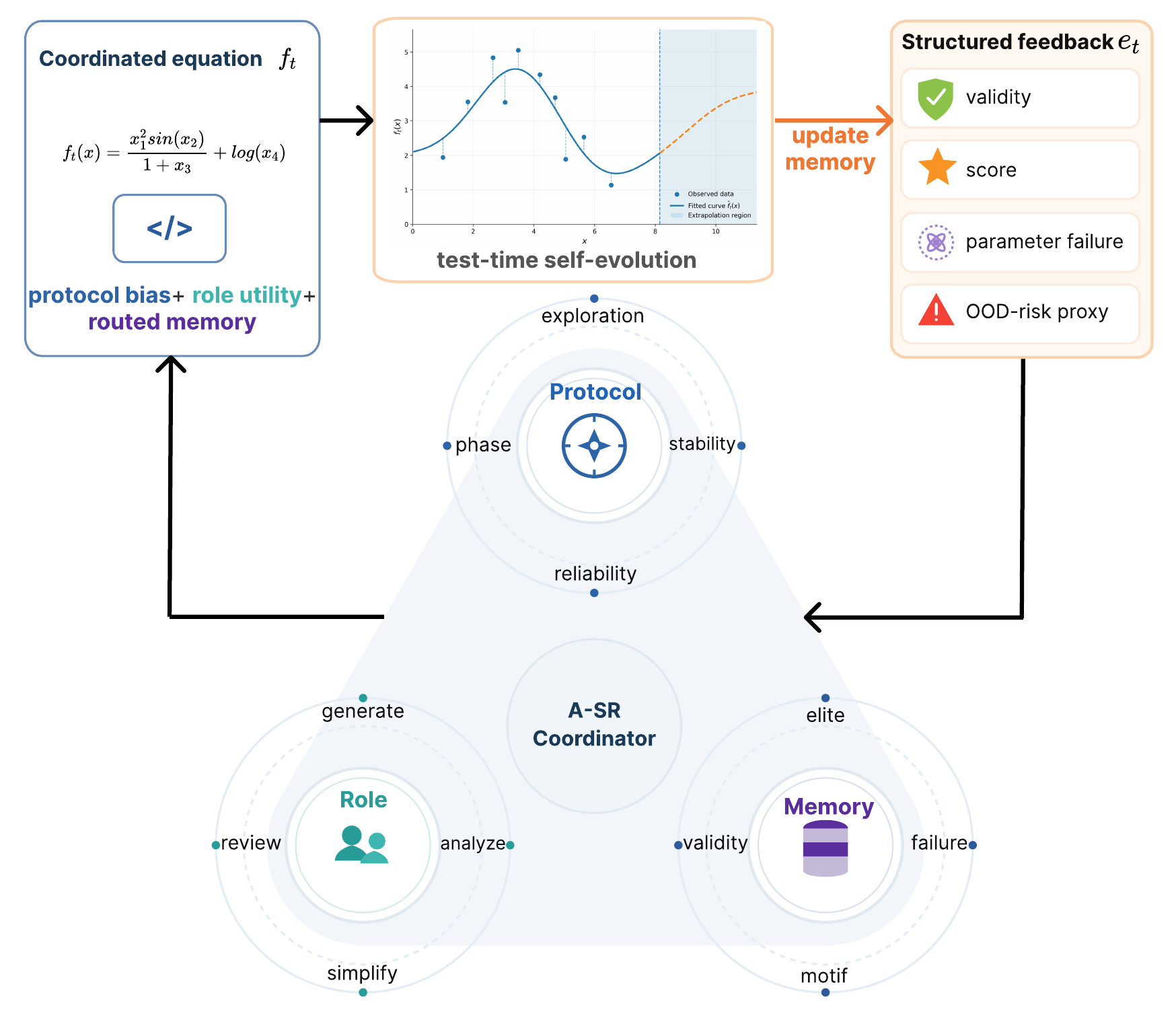}
\caption{Agentic feedback-driven coordination in A-SR. A-SR converts validity, improvement, failure, and complexity signals into coordination decisions over protocol, role, and routed memory, forming the context for the next equation.}
\label{fig:coordination_mechanism}
\end{figure}

Recent LLM-guided symbolic regression methods show that language models provide useful scientific and programmatic priors for equation discovery \citep{shojaee2024llmsr,grayeli2024lasr,shojaee2025llmsrbench}. By representing equations as executable programs, an LLM can propose equation skeletons while an evaluator fits continuous parameters and scores candidates \citep{chen2021evaluating,austin2021program,kwon2023vllm}. Nevertheless, many systems still rely on a unified proposal-evaluation loop: an LLM proposes a candidate, an evaluator returns a score, and subsequent prompts refine or resample formulas. Such a loop exposes limited process structure and often struggles to decide when to explore, repair invalidity, simplify, or validate. Our search logs show this mismatch directly: failures often persist because the next proposal receives evidence mismatched to the current failure mode, such as seeking novel structure when the dominant issue is invalidity, parameter instability, or redundant terms.


Recent agentic approaches separate proposal from navigation through planning, tools, reflection, memory, and structured interaction protocols \citep{yao2023react,schick2023toolformer,wu2023autogen,li2023camel}. In LLM-guided SR, Deliberate Evolution (DE) adapts symbolic edit operators, diagnostics, and reflective memory to steer how parent expressions are refined, mutated, crossed over, or regenerated \citep{pang2026deliberate}. However, operator-level adaptation decides what edit to apply; it does not fully determine what evidence the proposer sees. The same nominal operation can induce different search behavior when conditioned on elite motifs, failure traces, or validity diagnostics.

We argue that the locus of control should move from which edit to apply to which role sees which evidence. As illustrated in Figure~\ref{fig:coordination_mechanism}, A-SR implements this agentic control loop by turning evaluator feedback into coordinated choices over the protocol, active role-specialized agent, and routed memory view. A-SR makes the role--memory-view pair the unit of control: \emph{Generator}, \emph{Analyst}, \emph{Simplifier}, and \emph{Reviewer} agents receive different routed evidence under different search states. Within a run, A-SR selects a coordination protocol, adapts a lightweight non-stationary bandit-style role policy from evaluator-derived rewards, and routes state-conditioned memory without updating LLM parameters. Across runs, A-SR trajectories are distilled into supervised data for training open-source role-conditioned proposal priors, yielding A-SR-LoRA.

Empirically, this separation matters. On the four LSR-Synth scientific domains in LLM-SRBench, A-SR improves average Acc@0.01 over DE from 25.79\% to 48.30\% with Llama3.1-8B-Instruct, while A-SR-LoRA improves the corresponding Qwen3-4B-Instruct result from 24.58\% to 38.29\%. On four real-world scientific discovery tasks, A-SR obtains the best ID/OOD NMSE on 7 of 8 reported metrics. The gains do not come from role prompting alone: process diagnostics show that the full system maintains high validity without collapsing to a single role, and ablations show that fixed role rotation is insufficient without protocol selection, online role-policy adaptation, and routed memory.

Our contributions are:
\begin{itemize}
    \item \textbf{Formulation.} We recast LLM-guided symbolic regression as a control problem over role--memory-view pairs, driven by evaluator-observable failure modes rather than a single scalar reward.
    \item \textbf{Method.} We instantiate this formulation as A-SR, which self-evolves within a run through coordination protocol selection, evaluator-rewarded role adaptation, and state-routed memory without updating LLM parameters.
    \item \textbf{Distillation.} We further introduce A-SR-LoRA, which distills recorded A-SR trajectories into an open-weight role-conditioned proposal prior.
    \item \textbf{Results and analysis.} We evaluate static, online, and LoRA-distilled A-SR variants on symbolic regression benchmarks and provide diagnostics showing how validity, role usage, and memory routing change during search.
\end{itemize}

\section{Problem Formulation}

Given a dataset $\mathcal{D}=\{(\mathbf{x}_i, y_i)\}_{i=1}^{n}$, where $\mathbf{x}_i \in \mathbb{R}^d$ and $y_i \in \mathbb{R}$, symbolic regression seeks a closed-form expression $f(\mathbf{x}; \theta)$ with optimizable parameters $\theta$ that minimizes prediction error, $\min_{f,\theta} \mathcal{L}(f(\mathbf{x};\theta), y)$.

Each task is represented as a problem specification $P=(Q,\mathcal{X},\mathcal{D}_{\mathrm{train}},\mathcal{D}_{\mathrm{val}},\mathcal{D}_{\mathrm{ood}})$, where $Q$ is the fixed task description, $\mathcal{X}$ is the allowed variable set, and $\mathcal{D}$ contains numerical observations. In A-SR, candidate equations are represented as executable programs, following the program-skeleton view of LLM-guided SR \citep{shojaee2024llmsr}. At iteration $t$, the coordinator composes a role-conditioned prompt context $C_t=\mathrm{Compose}(Q,\mathcal{X},r_t,\mathcal{M}_t,e_{<t},B_t)$, where $r_t$ is the active role, $\mathcal{M}_t$ is routed process memory, $e_{<t}$ is recent evaluator feedback, and $B_t$ summarizes the current best candidate. Thus, $Q$ specifies the static discovery problem, whereas $C_t$ is the time-varying evidence view shown to the LLM. The LLM then outputs an equation program $f_t$, and a numerical evaluator executes $f_t$, fits continuous parameters, computes the score, and returns process feedback $e_t=(s_t,v_t,b_t,h_t,c_t)$, where $s_t$ is the score, $v_t$ records validity, $b_t$ indicates whether the candidate improves the current best, $h_t$ summarizes failure or parameter errors, and $c_t$ records complexity or diagnostic-state cues. We use lowercase $c_t$ for this feedback component and uppercase $C_t$ for the full prompt context. A-SR therefore operates on a role-conditioned process trace $\tau_t=\{(r_i,C_i,f_i,e_i)\}_{i=1}^{t}$ rather than only a population of expressions. This trace provides the evidence used for protocol selection, role-policy adaptation, and process-memory routing. A-SR does not change the SR objective above; instead, it uses $\tau_t$ to choose future contexts and roles so that subsequent candidates are more likely to reduce validation loss. The final output is the best valid expression discovered under the evaluator.

\begin{figure*}[t]
\centering
\includegraphics[width=\textwidth]{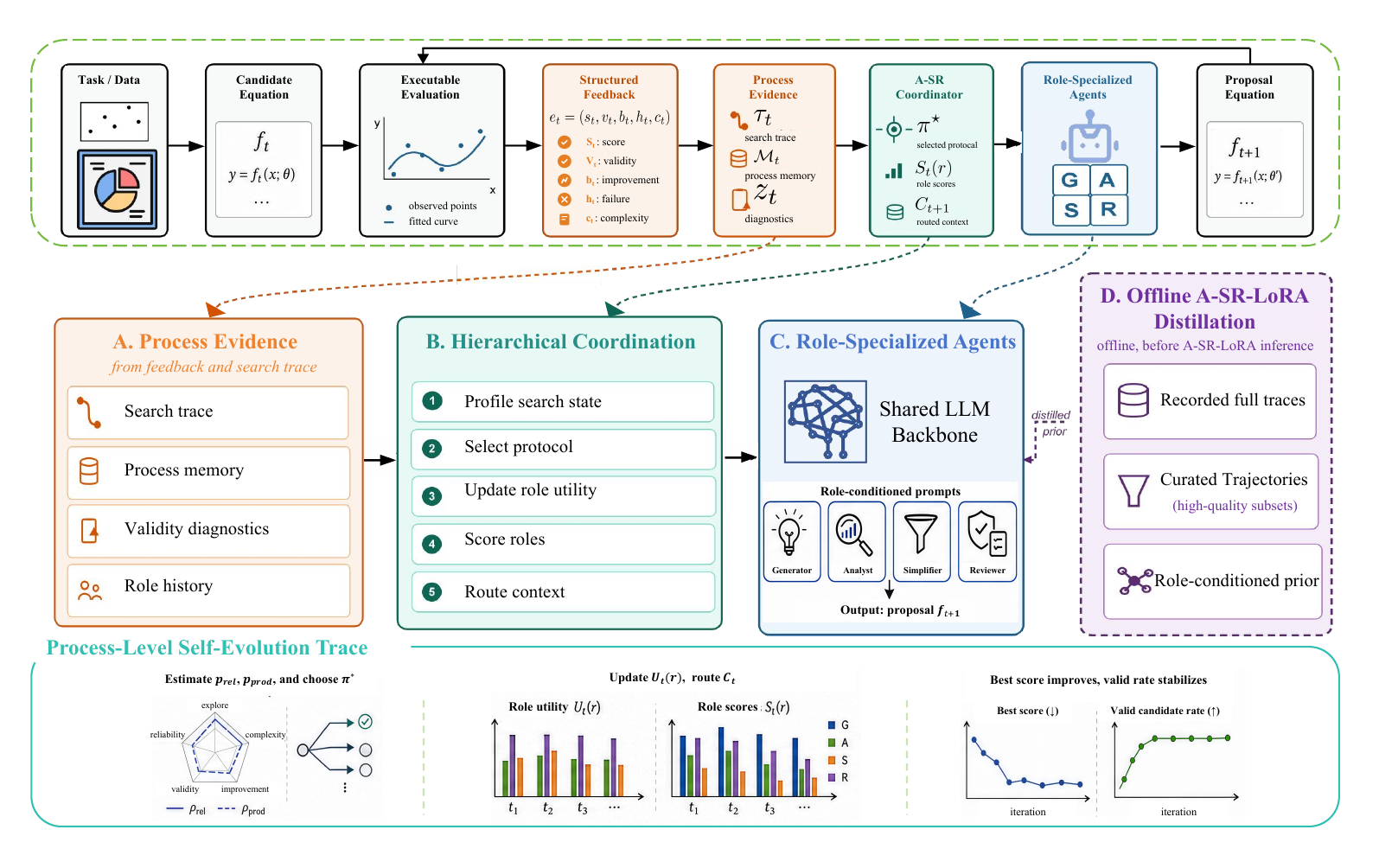}
\caption{Overall A-SR framework. A-SR converts executable evaluator feedback into process evidence, then uses a hierarchical coordinator to profile the search state, select a coordination protocol, update role utilities, score roles, and route a role-specific context. Role-specialized agents share an LLM backbone but receive different role instructions and memory views before proposing the next executable equation. The bottom trace shows test-time self-evolution of score, validity, protocol, utilities, and role scores, while the offline A-SR-LoRA path distills recorded trajectories into a role-conditioned proposal prior.}
\label{fig:framework}
\end{figure*}

\section{Method}

\subsection{Overview}

A-SR has four components, shown in Figure~\ref{fig:framework}: state-conditioned coordination, protocol selection, online role-policy adaptation, and role-aware process-memory routing. The main search is training-free with respect to the LLM backbone; self-evolution occurs by updating candidate archives, process memory, search state, role utilities, and protocol-conditioned contexts during test-time search. A-SR-Static disables online role-utility updates, while A-SR denotes the online variant unless otherwise specified. Given the trace after $t$ evaluated candidates, the next-step flow is $P,\tau_t \rightarrow C_{t+1} \rightarrow f_{t+1} \rightarrow e_{t+1} \rightarrow \tau_{t+1}$: the coordinator forms a routed context, the LLM proposes an executable formula, and the evaluator returns feedback.

\begin{table*}[t]
\centering
\small
\begin{tabular}{lcccccccccc}
\toprule
\textbf{Method} &
\multicolumn{2}{c}{\textbf{LSR-Transform}} &
\multicolumn{2}{c}{\textbf{Physics}} &
\multicolumn{2}{c}{\textbf{Material}} &
\multicolumn{2}{c}{\textbf{Chemistry}} &
\multicolumn{2}{c}{\textbf{Biology}} \\
\cmidrule(lr){2-3}
\cmidrule(lr){4-5}
\cmidrule(lr){6-7}
\cmidrule(lr){8-9}
\cmidrule(lr){10-11}
& {NMSE} & {Acc@0.01}
& {NMSE} & {Acc@0.01}
& {NMSE} & {Acc@0.01}
& {NMSE} & {Acc@0.01}
& {NMSE} & {Acc@0.01} \\
\midrule
\multicolumn{11}{c}{\textbf{Llama3.1-8B-Instruct}} \\
\midrule
LLMDirect & 2.95e-1 & 34.23 & 9.95e-3 & 0.00 & 8.16e-2 & 24.00 & 1.22e0 & 2.78 & 1.29e-1 & 4.17 \\
LLM-SR & 2.42e-1 & 34.23 & 3.00e-3 & 6.82 & 2.16e-1 & 60.00 & 5.24e-2 & 16.67 & 1.76e-2 & 12.50 \\
LASR & 2.62e-1 & 33.33 & 6.07e-3 & 9.09 & 9.47e-4 & 32.00 & 1.82e-3 & 8.33 & 6.40e-3 & 0.00 \\
SGA & 3.52e-1 & 0.90 & 1.55e-1 & 2.27 & 4.35e-2 & 12.00 & 4.58e-2 & 8.33 & 2.42e-1 & 0.00 \\
DE & 1.12e-1 & 36.04 & 1.01e-3 & 11.36 & 2.89e-4 & 64.00 & 4.16e-4 & 11.11 & 1.17e-2 & 16.67 \\
A-SR-Static
& \multicolumn{1}{c}{\textbf{1.01e-1}}
& \multicolumn{1}{c}{\underline{38.29}}
& \multicolumn{1}{c}{\textbf{8.46e-4}}
& \multicolumn{1}{c}{\underline{22.73}}
& \multicolumn{1}{c}{\underline{8.29e-6}}
& \multicolumn{1}{c}{\underline{80.00}}
& \multicolumn{1}{c}{\textbf{1.92e-5}}
& \multicolumn{1}{c}{\underline{26.39}}
& \multicolumn{1}{c}{\underline{8.66e-4}}
& \multicolumn{1}{c}{\underline{41.67}} \\
\textbf{A-SR}
& \multicolumn{1}{c}{\underline{1.09e-1}}
& \multicolumn{1}{c}{\textbf{40.54}}
& \multicolumn{1}{c}{\underline{9.03e-4}}
& \multicolumn{1}{c}{\textbf{27.27}}
& \multicolumn{1}{c}{\textbf{7.45e-6}}
& \multicolumn{1}{c}{\textbf{84.00}}
& \multicolumn{1}{c}{\underline{2.60e-5}}
& \multicolumn{1}{c}{\textbf{31.94}}
& \multicolumn{1}{c}{\textbf{5.38e-6}}
& \multicolumn{1}{c}{\textbf{50.00}} \\
\midrule
\multicolumn{11}{c}{\textbf{Qwen3-4B-Instruct-2507}} \\
\midrule
LLMDirect & 3.55e-1 & 24.32 & 5.46e-2 & 6.82 & 1.42e-3 & 52.00 & 2.66e-1 & 2.78 & 4.46e-2 & 8.33 \\
LLM-SR & 3.15e-1 & 26.13 & 2.51e-3 & 6.82 & 3.55e-3 & 44.00 & 3.36e-2 & 13.89 & 1.88e-2 & 12.50 \\
LASR & 1.83e-1 & 30.91 & 6.04e-3 & 6.82 & 6.21e-4 & 8.00 & 2.31e-3 & 2.78 & 9.56e-3 & 0.00 \\
SGA & 4.09e-1 & 19.81 & 1.04e-1 & 0.00 & 1.02e-2 & 16.00 & 1.61e-1 & 2.78 & 1.73e-1 & 4.17 \\
DE
& \multicolumn{1}{c}{\textbf{1.15e-1}}
& \multicolumn{1}{c}{\textbf{50.45}}
& \multicolumn{1}{c}{\textbf{4.37e-4}}
& 15.91
& 1.47e-4
& 56.00
& 1.88e-4
& 13.89
& 6.69e-3
& 12.50 \\
A-SR-Static
& \multicolumn{1}{c}{\underline{1.17e-1}}
& 32.43
& \multicolumn{1}{c}{\underline{5.31e-4}}
& 19.32
& \multicolumn{1}{c}{\textbf{3.54e-6}}
& 66.00
& 1.62e-5
& 33.33
& 1.68e-3
& 16.67 \\
\textbf{A-SR}
& 1.37e-1
& 33.33
& 6.68e-4
& \multicolumn{1}{c}{\underline{21.59}}
& \multicolumn{1}{c}{\underline{3.48e-5}}
& \multicolumn{1}{c}{\underline{68.00}}
& \multicolumn{1}{c}{\textbf{1.27e-5}}
& \multicolumn{1}{c}{\underline{34.72}}
& \multicolumn{1}{c}{\underline{1.25e-3}}
& \multicolumn{1}{c}{\underline{20.83}} \\
\textbf{A-SR-LoRA}
& 1.42e-1
& \multicolumn{1}{c}{\underline{34.23}}
& 1.91e-3
& \multicolumn{1}{c}{\textbf{22.73}}
& 3.00e-3
& \multicolumn{1}{c}{\textbf{70.00}}
& \multicolumn{1}{c}{\underline{1.33e-5}}
& \multicolumn{1}{c}{\textbf{37.50}}
& \multicolumn{1}{c}{\textbf{9.81e-4}}
& \multicolumn{1}{c}{\textbf{22.92}} \\
\bottomrule
\end{tabular}
\caption{Main LLM-SRBench results with Llama3.1-8B-Instruct and Qwen3-4B-Instruct-2507. We report NMSE ($\downarrow$) and Acc@0.01 ($\uparrow$, \%); bold and underlined values mark the best and second-best results per backbone.}
\label{tab:main_llmsrbench}
\end{table*}
\subsection{Role-Protocol Coordination}
A-SR instantiates four role-specialized agents as controller-selected behavioral modes over a shared LLM backbone. The roles are motivated by evaluator-observable failure modes in our traces: the \emph{Generator} proposes new equation structures, the \emph{Analyst} diagnoses missing terms and structural gaps, the \emph{Simplifier} compresses formulas toward compact and stable forms, and the \emph{Reviewer} checks validity, parameter usage, numerical stability, and extrapolation-risk proxies. The \emph{Reviewer} never accesses held-out OOD labels; its risk signals come only from executable feedback available during search, such as invalid programs, parameter-bound failures, expression complexity, and numerical instability. Unlike symbolic edit operators, these roles are coordinated according to discovery state rather than prescribing a fixed expression edit.

A-SR begins with an early profiling phase over the first $B_0$ evaluated candidates ($B_0=80$ in low-budget LLM-SRBench experiments). Roles follow a neutral fixed multi-agent schedule, online adaptation is disabled, and evaluator-derived process signals are collected.

We organize these signals into a reliability profile $\rho_{\mathrm{rel}}=(r_{\mathrm{invalid}}, r_{\mathrm{param}}, \Delta_{\mathrm{invalid}}, \Delta_{\mathrm{param}})$ and a productivity profile $\rho_{\mathrm{prod}}=(n_{\mathrm{best}}, n_{\mathrm{late}}, S_{\mathrm{best}}, \tau_{\mathrm{stag}})$. Here, $r_{\mathrm{invalid}}$ and $r_{\mathrm{param}}$ are invalid-candidate and parameter-failure rates, $\Delta_{\mathrm{invalid}}$ and $\Delta_{\mathrm{param}}$ are their recent changes, $n_{\mathrm{best}}$ and $n_{\mathrm{late}}$ count best-improving and late-window improvements, $S_{\mathrm{best}}$ is the current best score, and $\tau_{\mathrm{stag}}$ is steps since the last improvement.

After early profiling, A-SR selects a coordination protocol:
\begin{equation}
\begin{aligned}
    \pi^\star &=
    \mathrm{SelectProtocol}(\rho_{\mathrm{rel}}, \rho_{\mathrm{prod}}), \\
    \Pi &=
    \{\pi_{\mathrm{EGC}}, \pi_{\mathrm{RGC}}, \pi_{\mathrm{PGC}}, \pi_{\mathrm{SGC}}\}.
\end{aligned}
\end{equation}

The four protocols are \emph{Exploration-Guided Coordination} ($\pi_{\mathrm{EGC}}$), \emph{Reliability-Guided Coordination} ($\pi_{\mathrm{RGC}}$), \emph{Phase-Guided Coordination} ($\pi_{\mathrm{PGC}}$), and \emph{Stability-Guided Coordination} ($\pi_{\mathrm{SGC}}$). They are coordination archetypes derived from recurring trajectory regimes: productive exploration, reliability-dominated failure, stage-dependent exploration-to-consolidation, and unstable or stagnant search. They instantiate established search principles---exploration--exploitation control, repair and simplification, staged optimization, and adaptive exploration control---at the level of role and memory coordination \citep{fortin2012deap,cranmer2023pysr,madaan2023selfrefine,pang2026deliberate}. EGC encourages \emph{Generator}/\emph{Analyst} exploration; RGC emphasizes \emph{Reviewer}/\emph{Simplifier} behavior under invalidity, parameter failures, or excessive complexity; PGC shifts from exploration to regularization; and SGC increases exploration under stagnation while becoming conservative under unreliable proposals. The selector transparently maps reliability and productivity profiles to the protocol whose coordination bias best matches the observed regime. Thus, the novelty is not any individual search heuristic, but using evaluator-observable process signals to jointly decide which role acts and which evidence view that role receives.

\subsection{Online Adaptation and Memory}

Once a protocol is selected, A-SR performs online role-policy adaptation. Adaptation starts only after early profiling, so the protocol-selection profile is not affected by role-utility feedback. At iteration $t>B_0$, the active role is selected from $\mathcal{R}=\{\mathrm{Generator},\mathrm{Analyst},\mathrm{Simplifier},\mathrm{Reviewer}\}$.

A-SR maintains a utility $U_t(r)$ for each role $r$. After role $r_t$ proposes a candidate and the evaluator returns feedback, A-SR computes an evaluator-derived reward:
\begin{equation}
    R_t =
    \lambda_v \mathbb{I}_{\mathrm{valid}}
    +
    \lambda_b \mathbb{I}_{\mathrm{best}}
    -
    \lambda_i \mathbb{I}_{\mathrm{invalid}}
    -
    \lambda_p \mathbb{I}_{\mathrm{param}}
\end{equation}
where $\lambda_v,\lambda_b,\lambda_i,\lambda_p \geq 0$ are fixed across tasks and weight validity, best-improvement, invalidity, and parameter-failure signals.

The selected role utility is updated by
\begin{equation}
    U_t(r_t)
    =
    \mathrm{clip}
    \left(
    (1-\eta)U_{t-1}(r_t)
    +
    \eta R_t,
    -1,
    1
    \right),
\end{equation}
while utilities of unselected roles remain unchanged. Here $\eta \in (0,1]$ is the smoothing rate, and clipping keeps utilities bounded.

Role selection uses a protocol-conditioned score:
\begin{equation}
    S_t(r)
    =
    S_{\pi^\star}(r)
    +
    \alpha_{\pi^\star}
    g_{\pi^\star,t}(r)
    U_t(r),
\end{equation}
and selects
\begin{equation}
    r_t = \arg\max_{r \in \mathcal{R}} S_t(r).
\end{equation}

Here, $S_{\pi^\star}(r)$ is the protocol-induced base role score, $\alpha_{\pi^\star}$ controls adaptation strength, and $g_{\pi^\star,t}(r)$ is a protocol-conditioned gate. This adapts the role policy at test time without training the LLM backbone.

A-SR maintains process memory $\mathcal{M}_t=\{\mathcal{M}^{\mathrm{elite}}_t,\mathcal{M}^{\mathrm{fail}}_t,\mathcal{M}^{\mathrm{motif}}_t\}$, storing top valid formulas, recent failures, and recurring structural motifs. At each step, A-SR infers $z_t \in \{\mathrm{invalid},\mathrm{productive},\mathrm{stagnant},\mathrm{complex},\mathrm{mixed}\}$ from recent validity, improvement, parameter-failure, and complexity cues. The memory router exposes role- and state-specific blocks through $C_t=\mathrm{RouteMemory}(\mathcal{M}_t,z_t,r_t)$; detailed routing is expanded in the appendix.

For invalid-heavy states, the \emph{Reviewer} receives failure traces and validity diagnostics, while the \emph{Simplifier} receives repair cues. For valid-but-stagnant states, the \emph{Generator} and \emph{Analyst} receive motif memory for structural exploration. For complex-but-valid states, the \emph{Simplifier} receives compression cues and elite formula context.


\begin{figure}[t]
\centering
\includegraphics[width=\columnwidth]{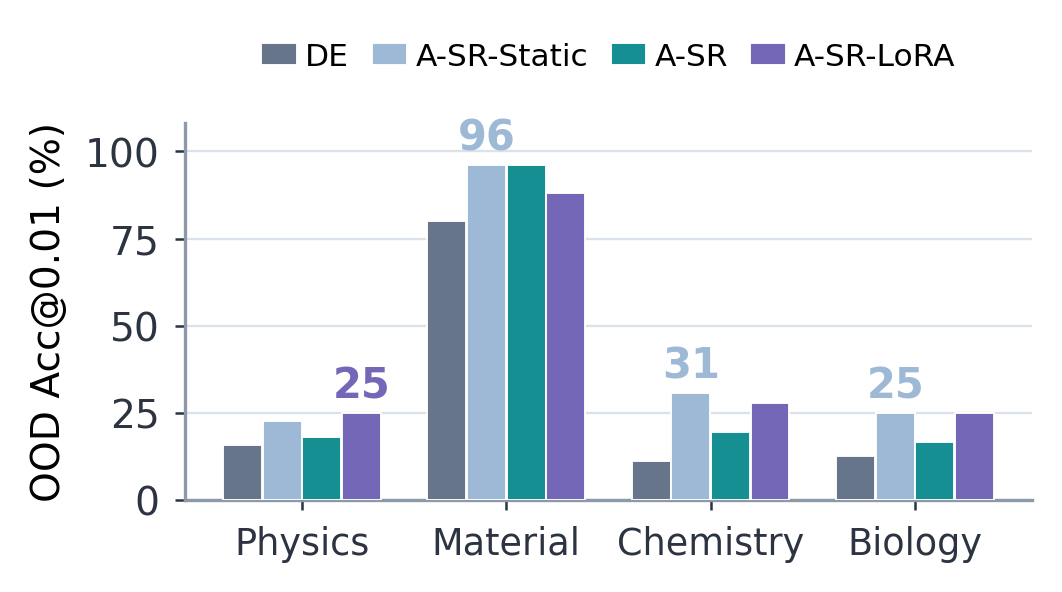}
\caption{OOD Acc@0.01 on the LSR-Synth scientific domains using Qwen3-4B-Instruct-2507. Full OOD NMSE and Acc@0.01 results are reported in Appendix Table~\ref{tab:ood_appendix}.}
\label{fig:ood_acc}
\end{figure}

\begin{table*}[t]
\centering
\small
\begin{tabular}{p{0.13\textwidth}*{8}{>{\centering\arraybackslash}p{0.08\textwidth}}}
\toprule
\textbf{Method} &
\multicolumn{2}{c}{\textbf{Oscillation 1}} &
\multicolumn{2}{c}{\textbf{Oscillation 2}} &
\multicolumn{2}{c}{\textbf{E. coli Growth}} &
\multicolumn{2}{c}{\textbf{Stress-Strain}} \\
\cmidrule(lr){2-3}
\cmidrule(lr){4-5}
\cmidrule(lr){6-7}
\cmidrule(lr){8-9}
& \makecell{ID\\NMSE} & \makecell{OOD\\NMSE}
& \makecell{ID\\NMSE} & \makecell{OOD\\NMSE}
& \makecell{ID\\NMSE} & \makecell{OOD\\NMSE}
& \makecell{ID\\NMSE} & \makecell{OOD\\NMSE} \\
\midrule
GPlearn & 1.55e-2 & 5.57e-1 & 7.55e-1 & 3.19e0 & 1.08e0 & 1.04e0 & 1.06e-1 & 4.09e-1 \\
NeSymReS & 4.70e-3 & 5.38e-1 & 2.49e-1 & 6.47e-1 & NA & NA & 7.93e-1 & 6.38e-1 \\
E2E & 8.20e-3 & 3.72e-1 & 1.40e-1 & 1.91e-1 & 6.32e-1 & 1.45e0 & 2.26e-1 & 5.87e-1 \\
DSR & 8.70e-3 & 2.45e-1 & 5.80e-2 & 1.95e-1 & 9.45e-1 & 2.43e0 & 3.33e-1 & 1.11e0 \\
uDSR & 3.00e-4 & 7.00e-4 & 3.20e-3 & 1.50e-3 & 3.32e-1 & 5.46e0 & 5.02e-2 & 1.76e-1 \\
PySR & 9.00e-4 & 3.11e-1 & 2.00e-4 & 9.80e-3 & 3.76e-2 & 1.01e0 & 3.31e-2 & 1.30e-1 \\
LLM-SR & \underline{4.65e-7} & \underline{5.00e-4} & \underline{2.12e-7} & \underline{3.81e-5} & \underline{2.14e-2} & \underline{2.64e-2} & \underline{2.10e-2} & \textbf{5.16e-2} \\
\textbf{A-SR} & \textbf{1.80e-7} & \textbf{1.82e-4} & \textbf{1.10e-7} & \textbf{4.54e-7} & \textbf{9.00e-3} & \textbf{1.46e-2} & \textbf{2.01e-2} & \underline{5.44e-2} \\
\bottomrule
\end{tabular}
\caption{Real-world scientific discovery results using GPT-3.5. We report ID and OOD NMSE ($\downarrow$); published baseline values are taken from LLM-SR when available. DE is not included because it is not open-sourced and does not provide detailed results.}
\label{tab:main_realworld}
\end{table*}

\subsection{Trajectory Distillation}

Although A-SR adapts at test time without updating LLM parameters, its trajectories provide structured supervision for training an open-source role-conditioned proposal prior. We apply trajectory distillation to Qwen3-4B-Instruct: Llama3.1-8B experiments use the training-free coordinator, while Qwen experiments additionally evaluate A-SR-LoRA.

With full trajectory logging, each record contains the task specification, active role, process state, routed memory context, generated formula, validity, score, improvement indicator, and failure type. We retain valid improvements, top valid candidates, successful repairs, compact simplifications, and role-consistent proposals as supervised examples for a role-conditioned agent prior:
\begin{equation}
    p_{\phi}
    \left(
    f_t
    \mid
    P,
    r_t,
    z_t,
    C_t
    \right),
\end{equation}
where $P$ is the problem specification, $r_t$ is the role, $z_t$ is the process state, and $C_t$ is routed context.

A-SR-LoRA keeps the A-SR coordination loop but replaces the proposal backbone with the LoRA-distilled model. The evaluator, protocol selector, role policy, and memory router remain unchanged during LoRA inference. Process diagnostics are reported when full coordinator traces are logged.

\section{Experiments}

\subsection{Experimental Setup}

We evaluate A-SR in two settings. First, we use LLM-SRBench, an open benchmark for LLM-guided symbolic regression that contains LSR-Transform and LSR-Synth tasks spanning physics, material science, chemistry, and biology \citep{shojaee2025llmsrbench}. Second, we evaluate on the four real-world scientific discovery tasks introduced by LLM-SR: Oscillation 1, Oscillation 2, E. coli Growth, and Stress-Strain \citep{shojaee2024llmsr}; the Stress-Strain task is built from real experimental tensile measurements on aluminum 6061-T651 \citep{aakash2019stressstrain}.

For LLM-SRBench, we compare against direct LLM generation, LLM-SR-style proposal search, LASR, SGA, and Deliberate Evolution when reported under matched backbones \citep{shojaee2024llmsr,grayeli2024lasr,pang2026deliberate}. We evaluate A-SR with Llama3.1-8B-Instruct and Qwen3-4B-Instruct-2507 \citep{dubey2024llama3,yang2025qwen3}. We also report two A-SR variants when available: A-SR-Static retains protocol selection and role-aware memory routing but disables online role-utility adaptation; A-SR is the online variant with evaluator-rewarded role-utility updates; A-SR-LoRA plugs a trajectory-distilled Qwen LoRA backbone back into the same A-SR coordination loop. For the real-world tasks, we compare against the reported GPlearn, NeSymReS, E2E, DSR, uDSR, PySR, and LLM-SR baselines \citep{biggio2021neural,kamienny2022end,petersen2021deep,landajuela2022unified,cranmer2023pysr,shojaee2024llmsr}. We report DE only where the benchmark and settings are comparable; for the four LLM-SR real-world tasks, DE does not provide a complete reproducible result table.

Unless otherwise stated, we report normalized mean squared error (NMSE) and Acc@0.01. Following the LLM-SRBench protocol, Acc@0.01 measures the fraction of tasks whose predictions satisfy the 1\% relative-error tolerance after discarding the worst 5\% predictions. For A-SR, main LLM-SRBench values average train and validation performance, while OOD performance is reported separately in the appendix.

\begin{figure*}[t]
\centering
\includegraphics[width=\textwidth]{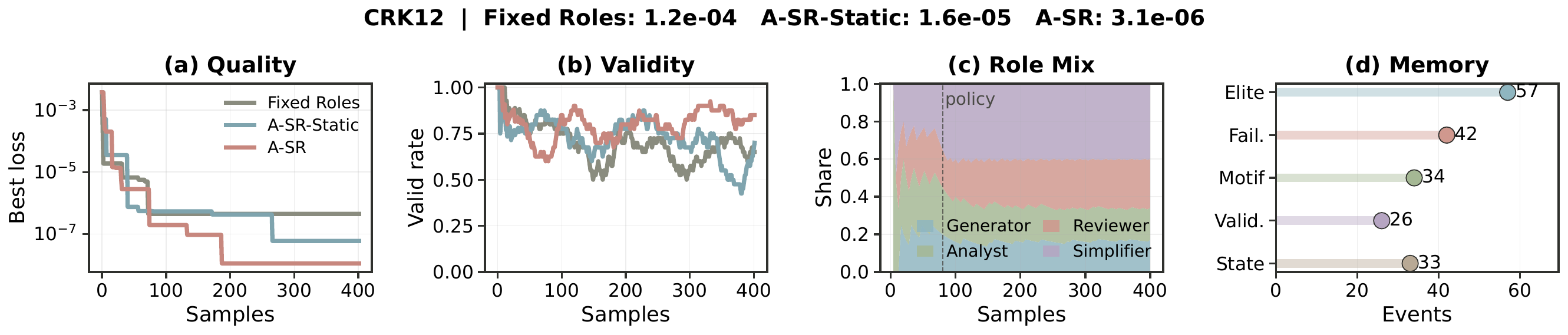}
\caption{Search dynamics on Chemistry task CRK12. A-SR improves best-so-far loss while maintaining high validity, non-collapsed role usage, and routed process memory.}
\label{fig:best_score_trajectories}
\end{figure*}
\subsection{Main Results on LLM-SRBench}

Table~\ref{tab:main_llmsrbench} reports the main LLM-SRBench results. With Llama3.1-8B-Instruct, A-SR-Static achieves the best LSR-Transform NMSE, while online A-SR obtains the best LSR-Transform Acc@0.01 and the best Acc@0.01 on Physics, Material, Chemistry, and Biology. Online A-SR also achieves the lowest NMSE on Material and Biology, and the second-best NMSE on LSR-Transform, Physics, and Chemistry. The key finding is a division of behavior: the static coordinator is strong for broad transformed-equation search, while online role-policy adaptation improves solved-task reliability on scientific-domain subsets.

With Qwen3-4B-Instruct-2507, the comparison is more mixed but still informative. A-SR-Static achieves strong NMSE aggregates on LSR-Transform and Material, while DE remains strongest on Physics NMSE. A-SR-LoRA improves solved-task accuracy on several scientific-domain subsets. The key finding is that the three variants improve different parts of the discovery process: static coordination is useful for broad expression recovery, online adaptation improves reliability when search states change over time, and trajectory distillation strengthens the role-conditioned proposal prior for smaller open-source backbones.

Figure~\ref{fig:ood_acc} reports Qwen3-4B-Instruct-2507 OOD Acc@0.01 on the LSR-Synth scientific domains, and Appendix Table~\ref{tab:ood_appendix} reports the full OOD NMSE and Acc@0.01 results using held-out extrapolation splits where available. We emphasize OOD Acc@0.01 in the main text because OOD NMSE can be dominated by a small number of catastrophic extrapolation failures. The Qwen results show that A-SR variants improve solved-task extrapolation reliability on Material, Chemistry, and Biology, while Physics remains volatile due to oscillatory structure and phase-sensitive extrapolation.

\subsection{Real-World Scientific Discovery Tasks}

Table~\ref{tab:main_realworld} reports results on the four LLM-SR real-world scientific discovery tasks using GPT-3.5. A-SR obtains the best ID and OOD NMSE on Oscillation 1, Oscillation 2, and E. coli Growth, and the best ID NMSE on Stress-Strain. On Stress-Strain OOD, A-SR is second-best behind LLM-SR by a small margin. The key finding is that A-SR improves both oscillator and growth discovery and remains competitive on the noisier material-response task: hierarchical coordination improves the reliability of finding valid, high-quality formulas across tasks, while individual material extrapolation cases can still depend on the exact functional family discovered.

\subsection{Search Dynamics}

Beyond final error, we analyze how methods use the sample budget and search state. Figure~\ref{fig:best_score_trajectories} shows a representative CRK12 trajectory comparing fixed role rotation, A-SR-Static, and online A-SR. In this representative run, online A-SR reaches a stronger final candidate while maintaining non-collapsed role usage. The same trace shows that \emph{Generator}, \emph{Analyst}, \emph{Reviewer}, and \emph{Simplifier} remain active, and that different memory types are routed into the search across process states. 

\begin{figure}[t]
\centering
\includegraphics[width=\columnwidth]{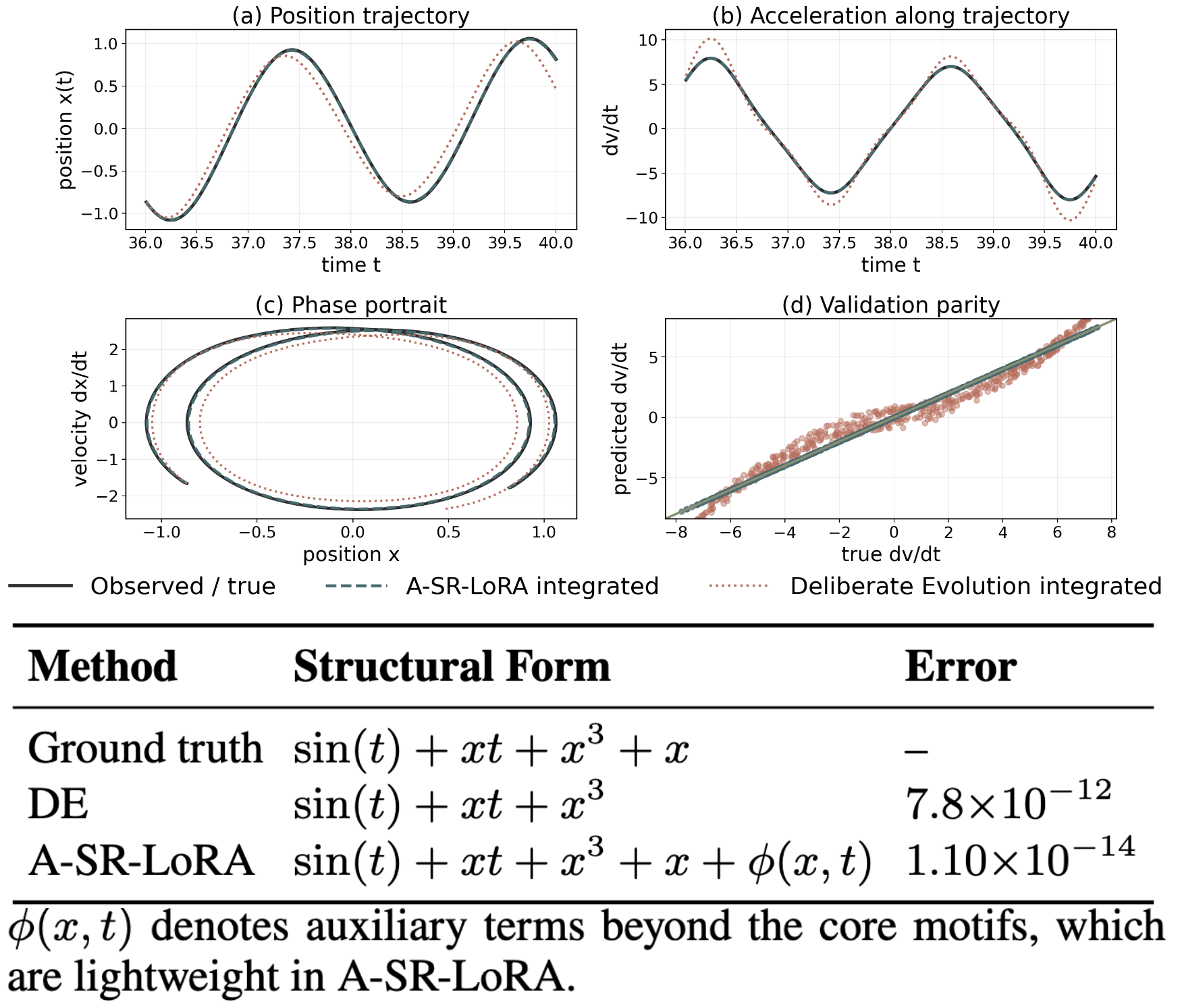}
\caption{Formula-level case study. A-SR recovers a scientifically meaningful expression with strong ID/OOD fit; additional examples are reported in Appendix Table~\ref{tab:case_formula_comparison}.}
\label{fig:case_study_example}
\end{figure}

These examples complement aggregate benchmark metrics. In the transformed scientific equations, A-SR often recovers algebraically equivalent forms up to fitted constants, showing that it can identify exact symbolic skeletons. In the Biology case, the discovered expression is not algebraically identical to the reference expression, but it captures a growth--saturation--decay mechanism and transfers well OOD, illustrating mechanism-level discovery.

\subsection{Ablations}

We ablate the components that distinguish A-SR from unified LLM-SR loops and fixed multi-agent prompting on a mixed 45-task subset of LLM-SRBench. The ablations remove role-aware memory routing, adaptive protocol selection, online role-policy adaptation, or the full hierarchical coordination stack. This design asks whether A-SR benefits from a single component or from the interaction between protocol-level coordination, role-level online adaptation, and state-conditioned memory exposure.

\begin{table}[t]
\centering
\footnotesize
\begin{tabular}{lcc}
\toprule
\textbf{Setting} & \textbf{Val Acc} & \textbf{OOD Acc} \\
\midrule
fixed role rotation & 24.44 & 24.44 \\
w/o routed memory & 26.67 & 24.44 \\
w/o adaptive protocol selection & 26.67 & \underline{33.33} \\
A-SR-Static & \underline{28.89} & \underline{33.33} \\
\textbf{A-SR online} & \textbf{31.11} & \textbf{35.56} \\
\bottomrule
\end{tabular}
\caption{Component ablation on a mixed 45-task LLM-SRBench subset with Llama3.1-8B-Instruct. Full NMSE results are reported in Appendix Table~\ref{tab:appendix_mixed45_ablation}.}
\label{tab:ablation}
\end{table}

Table~\ref{tab:ablation} shows that A-SR online achieves the highest validation Acc@0.01 and OOD Acc@0.01 on the mixed subset. We emphasize Acc@0.01 in the main-text ablation because mean NMSE in symbolic regression can be dominated by a few catastrophic extrapolation failures. Fixed role rotation performs substantially worse, showing that simply exposing the LLM to multiple role prompts is insufficient. Removing routed memory or adaptive protocol selection also reduces solved-task reliability, indicating that the gains come from coordinating what information each role receives and when each collaboration mode is used. Full NMSE values are reported in Appendix Table~\ref{tab:appendix_mixed45_ablation}.

The ablation settings isolate different levels of the coordination stack. Fixed role rotation keeps the four role prompts but removes protocol routing, online utility updates, and routed memory. The variant without routed memory keeps the coordinator but removes state-conditioned memory exposure. The variant without adaptive protocol selection disables the early profile-to-protocol routing and uses a fixed coordination protocol. A-SR-Static retains protocol selection and routed memory but disables online role-policy adaptation, while A-SR online is the full test-time system.

\subsection{Process and Case Study Analysis}

We further analyze A-SR through process-level diagnostics and formula-level examples. Figure~\ref{fig:best_score_trajectories} shows that online A-SR improves best-so-far loss while maintaining a high valid-candidate rate and a non-collapsed role usage. Figure~\ref{fig:case_study_example} complements this process view by comparing discovered symbolic structure rather than only aggregate error: A-SR recovers core scientific motifs with strong ID/OOD fit, whereas weaker formulas may obtain low validation error through auxiliary terms that are less aligned with the reference structure. Additional diagnostic-state, memory-routing, and formula-level analyses are provided in the appendix. These diagnostics are useful because a symbolic regression system can fail in qualitatively different ways: it may repeatedly emit invalid programs, converge to overly complex formulas, stagnate after finding a partial structure, or fit ID data while extrapolating poorly.

\section{Related Work}

\paragraph{Symbolic and LLM-guided regression.}
Symbolic regression has been studied through scientific law discovery, evolutionary search, reinforcement learning, and neural sequence models \citep{schmidt2009distilling,sun2023symbolicphysics,cranmer2023pysr,petersen2021deep,landajuela2022unified,biggio2021neural,kamienny2022end}. Classical SR research has also emphasized benchmark design, grammar or operator constraints, and the tension between compactness, accuracy, and extrapolation \citep{keijzer2003improving,vladislavleva2009order,orzechowski2018we,fortin2012deap,virgolin2021improving}. Benchmark studies such as SRBench and SRSD emphasize the difficulty of comparing SR systems fairly across equation families and data regimes \citep{lacava2021contemporary,matsubara2024rethinking}. LLM-SR reframes equation discovery as programming with LLM-generated equation skeletons and numerical parameter fitting \citep{shojaee2024llmsr}, while LLM-SRBench provides a benchmark designed for LLM-integrated SR beyond rote recitation \citep{shojaee2025llmsrbench}. A-SR adopts the executable-program formulation but makes evaluator feedback a coordination signal for roles, protocols, and memory, rather than only a scalar score for iterative proposal refinement.

\paragraph{Agentic and evaluator-in-the-loop search.}
Recent agentic systems use reasoning traces, self-consistency, structured deliberation, reflection, memory, tools, and executable evaluators to improve problem solving and program discovery \citep{wei2022chainofthought,wang2023selfconsistency,yao2023treeofthoughts,madaan2023selfrefine,shinn2023reflexion,romeraparedes2023funsearch}. Multi-agent and tool-augmented LLM systems further show that decomposing a task across roles, tools, or interaction protocols can improve robustness and inspectability in complex workflows \citep{li2023camel,park2023generative,qin2023toolllm,liu2023agentbench,wu2023autogen}. DE is the closest prior work in LLM-guided SR: it separates proposal from navigation through adaptive symbolic edit operators, analytical tools, and reflective memory \citep{pang2026deliberate}. A-SR is organized at a different control level: instead of selecting expression-edit operators around a parent formula, A-SR coordinates which scientific role acts and which memory view that role receives.

\section{Limitations}

A-SR uses explicit coordinator rules and lightweight online utilities rather than a learned neural controller. This design improves transparency and compatibility with both open-source and closed-source LLM backbones, but the protocol selector may require recalibration for substantially different scientific domains. Our results also show that different variants excel under different metrics: A-SR-Static can be stronger on some NMSE aggregates, online A-SR improves solved-task accuracy, and A-SR-LoRA benefits scientific-domain discovery on Qwen. Future work may learn protocol selection from larger meta-SR task banks or combine trajectory distillation with RL-style controller training.

\section{Conclusion}

We presented A-SR, a self-evolving agentic framework for symbolic regression. A-SR coordinates role-specialized LLM agents through coordination protocol selection, online role-policy adaptation, and role-aware process memory routing. Across static, online, and LoRA-distilled variants, A-SR improves scientific-domain success rates and provides an interpretable view of how evaluator feedback shapes discovery. By shifting LLM-guided SR from unified proposal loops and edit-centric adaptation toward hierarchical agent coordination, A-SR provides a flexible framework for reliable scientific equation discovery.





\bibliography{aaai2027}

\clearpage
\appendix

\begin{table*}[t]
\centering
\small
\setlength{\tabcolsep}{4pt}
\renewcommand{\arraystretch}{1.08}
\begin{tabular*}{\textwidth}{@{\extracolsep{\fill}}p{0.14\textwidth}p{0.13\textwidth}p{0.31\textwidth}p{0.34\textwidth}@{}}
\toprule
\textbf{Task} & \textbf{Method} & \textbf{Ground Truth} & \textbf{Equivalent Fitted Output} \\
\midrule
$I.10.7\_1\_0$
& A-SR online
& $-c\sqrt{1-m_0^2/m^2}$
& $-0.99999998\,c(1-(m_0/m)^2)^{0.5}$ \\

$II.27.16\_2\_0$
& A-SR online
& $-\sqrt{\mathrm{flux}/(c\epsilon)}$
& $-0.99999999\,\sqrt{\mathrm{flux}}/(\sqrt{\epsilon}\sqrt{c})$ \\

$II.34.2\_1\_0$
& A-SR online
& $2\,\mathrm{mom}/(qr)$
& $2.00000000\,\mathrm{mom}/(qr)$ \\

$II.34.2\_1\_0$
& A-SR-LoRA
& $2\,\mathrm{mom}/(qr)$
& $1.99999999\,\mathrm{mom}/(qr)$ \\

$I.44.4\_4\_0$
& A-SR online
& $V_1\exp(E_n/(T k_b n))$
& $0.99999807\,V_1\exp(E_n/(n k_b T))$ \\

$II.8.31\_1\_0$
& A-SR-Static
& $-\sqrt{2}\sqrt{E_{\mathrm{den}}/\epsilon}$
& $-1.41421359\,\sqrt{E_{\mathrm{den}}/\epsilon}$ \\
\bottomrule
\end{tabular*}
\caption{Formula-level case studies for algebraic recovery. A-SR recovers expressions that are algebraically equivalent to the reference after fitting scalar constants.}
\label{tab:case_formula_comparison}
\end{table*}

\section{Extended Related Work}

\paragraph{Classical and neural symbolic regression.}
Modern genetic-programming and evolutionary SR systems provide efficient search baselines for interpretable expression discovery \citep{fortin2012deap,cranmer2023pysr}. Scientific equation discovery work further showed that symbolic regression can recover compact physical laws from experimental data \citep{schmidt2009distilling,sun2023symbolicphysics}. Modern SR systems include efficient evolutionary implementations such as PySR, reinforcement-learning approaches such as DSR and uDSR, and neural sequence models such as NeSymReS and E2E \citep{cranmer2023pysr,petersen2021deep,landajuela2022unified,biggio2021neural,kamienny2022end}. Transformer architectures provide a common backbone for several neural symbolic-regression systems \citep{vaswani2017attention}. SRBench and SRSD provide standardized views of the strengths and weaknesses of these methods across classical benchmark suites \citep{lacava2021contemporary,matsubara2024rethinking}. These works motivate reporting both error and success rate, since a method can achieve low average error while solving relatively few tasks robustly.

\paragraph{LLM-integrated and agentic symbolic regression.}
LLM-SR introduces an executable-program view in which the LLM proposes equation skeletons and an external optimizer fits continuous constants \citep{shojaee2024llmsr}. LLM-SRBench extends this direction by constructing transformed and synthetic scientific tasks intended to reduce direct memorization by LLMs \citep{shojaee2025llmsrbench}. LASR explores learned concept libraries for symbolic regression \citep{grayeli2024lasr}. DE is the closest agentic SR method: it adds adaptive symbolic edit operators, diagnostic tools, and reflective memory to LLM-guided SR \citep{pang2026deliberate}. A-SR differs by coordinating scientific roles, coordination protocols, online role utilities, and state-conditioned process memory.

\section{Task and Dataset Details}

\paragraph{Benchmark tasks.}
Our benchmark evaluation follows the task organization used by recent LLM-guided symbolic regression studies, where equation discovery is evaluated not only on random symbolic expressions but also on scientific families with structured variables and extrapolation splits. We use LLM-SRBench as the main large-scale open benchmark because it contains transformed symbolic-regression problems and domain-oriented scientific tasks, allowing us to evaluate both broad expression recovery and scientific generalization. In the main paper, we group the tasks into LSR-Transform and LSR-Synth subsets, with the latter covering Physics, Material Science, Chemistry, and Biology.

\paragraph{Real-world scientific tasks.}
In addition to LLM-SRBench, we retain the four real-world scientific discovery tasks introduced in LLM-SR: two oscillator systems, an E. coli growth task, and a stress-strain relation. These tasks are useful because they test different failure modes of LLM-guided SR. Oscillator tasks stress extrapolation and phase-sensitive dynamics; growth dynamics tests saturation and nonlinear biological mechanisms; and stress-strain data tests material-response modeling under noisy and parameter-sensitive regimes.

\paragraph{Input-output format.}
Each task is provided to A-SR as a task description, a list of available variables, an executable equation-program template, and train/validation/OOD data splits when available. The LLM does not directly observe target equations. Instead, each role proposes executable candidate functions over the allowed variables. The evaluator then fits free constants and computes the corresponding metrics. This format follows the programmatic equation-discovery setting while making all role-specific proposals comparable under the same evaluator.
\begin{table}[t]
\centering
\small
\setlength{\tabcolsep}{4pt}
\renewcommand{\arraystretch}{1.08}
\begin{tabular*}{\columnwidth}{@{\extracolsep{\fill}}lll@{}}
\toprule
\textbf{State} & \textbf{Roles} & \textbf{Context} \\
\midrule
Invalid-heavy & Rev./Simp. & Failure, validity \\
Stagnant-valid & Gen./Anal. & Elite, motif \\
Complex-valid & Simp./Rev. & Compression cues \\
Productive & Gen./Anal. & Best motifs \\
Mixed & Policy-dependent & Balanced context \\
\bottomrule
\end{tabular*}
\caption{Qualitative state-conditioned memory routing.}
\label{tab:appendix_memory_routing}
\end{table}
\paragraph{Train, validation, and OOD splits.}
For LLM-SRBench experiments, we report train/validation aggregate performance in the main table and reserve OOD metrics for separate reporting, since OOD splits are not uniformly defined across all benchmark groups. For the real-world scientific tasks, we report both ID and OOD NMSE when the split is available. This separation avoids mixing interpolation accuracy and extrapolation robustness into a single number, which is especially important for scientific equation discovery.

\section{Benchmark and Memorization Limitations}

\paragraph{Why transformed and scientific benchmarks are needed.}
Feynman-style symbolic-regression benchmarks remain useful for testing algebraic recovery, but they are not by themselves sufficient evidence of scientific discovery with modern LLMs. Many canonical equations, variable names, and textbook derivations are likely to appear in pretraining corpora, so exact recovery on a familiar expression can mix genuine search behavior with memorized scientific form. This concern is especially important for LLM-guided SR because the model can exploit semantic cues in variable names, units, or problem descriptions even when it does not directly observe the target equation.

\paragraph{Benchmark choice in this paper.}
We therefore use LLM-SRBench as the main large-scale benchmark. Its transformed-equation tasks reduce the chance that a model simply emits a memorized textbook formula, while its scientific synthetic domains evaluate structured extrapolation across physics, material science, chemistry, and biology. The real-world LLM-SR tasks provide an additional check in settings where the target behavior is tied to noisy or experimentally motivated scientific phenomena. These benchmarks still do not eliminate memorization risk: transformed equations can preserve recognizable algebraic motifs, and scientific task descriptions can still provide useful prior information. For this reason, we report numerical accuracy, OOD behavior, formula structure, and search dynamics rather than relying only on exact symbolic match.

\paragraph{Interpretation of formula recovery.}
Our formula-level examples should be read as structural evidence rather than as a claim that A-SR never benefits from the LLM's scientific prior. A-SR intentionally uses the LLM as a proposal prior, but the main question is whether evaluator-guided coordination can turn that prior into reliable executable search. We therefore emphasize cases where the final expression is reached through valid intermediate candidates, role-specific repairs, and memory routing, and we report Acc@0.01 and OOD metrics to separate broad solved-task reliability from isolated exact-recovery examples.

\section{Additional Formulation Details}

\begin{algorithm}[t]
\footnotesize
\caption{A-SR Search Procedure}
\label{alg:asr}
\begin{algorithmic}[1]
\REQUIRE Dataset $\mathcal{D}$, problem specification $P$, LLM $M$, budget $B$, profiling budget $B_0$
\STATE Initialize process memory $\mathcal{M}_0$, role utilities $U_0(r)=0$, best score $S_{\mathrm{best}}=-\infty$
\FOR{$t=1$ to $B$}
    \IF{$t \leq B_0$}
        \STATE Select role $r_t$ using neutral multi-agent schedule
    \ELSE
        \IF{$t=B_0+1$}
            \STATE Build profile $(\rho_{\mathrm{rel}},\rho_{\mathrm{prod}})$ and select protocol $\pi^\star$
        \ENDIF
        \STATE Select role $r_t$ using protocol-conditioned role policy and online utility bias
    \ENDIF
    \STATE Infer process state $z_t$ and route memory context $C_t$
    \STATE Query LLM $M$ with $(P,r_t,C_t)$ to generate candidate $f_t$
    \STATE Evaluate $f_t$ and obtain score, validity, and failure signals
    \STATE Update process memory $\mathcal{M}_{t+1}$
    \IF{$t > B_0$ and online adaptation is enabled}
        \STATE Update role utility $U_t(r_t)$ from evaluator feedback
    \ENDIF
\ENDFOR
\RETURN Best valid equation found
\end{algorithmic}
\end{algorithm}

\paragraph{Executable program search.}
Algorithm~\ref{alg:asr} summarizes the full A-SR search procedure, including profiling, protocol selection, role-conditioned generation, evaluator feedback, memory updates, and online role adaptation. For each task, the LLM outputs an executable equation program rather than a raw expression string. The evaluator parses and executes the program, fits free constants, and rejects candidates that violate syntactic, numerical, or variable-usage constraints. We write this evaluator as
\begin{equation}
    e_t = \mathrm{Eval}(f_t;\mathcal{D}_{\mathrm{train}},\mathcal{D}_{\mathrm{val}}),
\end{equation}
where $f_t$ is the generated program and $e_t$ is the structured feedback tuple used by the coordinator. This abstraction separates symbolic structure proposal from continuous parameter fitting, which is handled by a numerical optimizer.

\paragraph{Process trace.}
A-SR maintains a role-conditioned process trace
\begin{equation}
    \tau_t=\{(r_i,C_i,f_i,e_i)\}_{i=1}^{t}.
\end{equation}
Here $r_i$ is the active role, $C_i$ is the routed context shown to the LLM, $f_i$ is the candidate formula program, and $e_i$ is evaluator feedback. This trace is the main state of the search. It differs from a population-only view because the coordinator reasons not only about which formulas scored well, but also about which role acted, what information was routed, and what failure mode appeared.

\begin{table*}[t]
\centering
\small
\setlength{\tabcolsep}{4pt}
\renewcommand{\arraystretch}{1.08}
\begin{tabular*}{\textwidth}{@{\extracolsep{\fill}}p{0.18\textwidth}p{0.76\textwidth}@{}}
\toprule
\textbf{Block} & \textbf{Example content} \\
\midrule
Task specification &
Discover an equation program for target $y$ using variables $x$, $v$, and $t$. Use only allowed variables and evaluator-supported operations. \\
Current best &
Best valid candidate: $-p_0 v - p_1 x$. Validation NMSE: $1.7{\times}10^{-2}$. Recent validation residuals suggest missing nonlinear stiffness. \\
Recent feedback &
Recent candidates are valid but stagnant. Parameter fitting is stable; improvements require new structure rather than additional constants. \\
Routed memory &
Elite memory: damping terms in $v$ are useful. Motif memory: $x^3$ and $v\sin(x)$ appeared in improving candidates. Failure memory: nested trigonometric forms caused unstable fitting. \\
Role instruction &
Active role: \emph{Analyst}. Propose one executable candidate that changes the symbolic structure while keeping the expression numerically stable. \\
Output constraint &
Return only a Python-style equation body over the allowed variables and free parameters, e.g., \texttt{return -p0*v - p1*x + p2*v*sin(x)}. \\
\bottomrule
\end{tabular*}
\caption{Condensed A-SR prompt example. The example illustrates the information blocks exposed to a role-conditioned LLM call; the candidate equation body is generated by the model and evaluated by the same executable-program evaluator used for all A-SR variants.}
\label{tab:appendix_prompt_example}
\end{table*}

\paragraph{Evaluator feedback.}
The feedback tuple is
\begin{equation}
    e_t=(s_t,v_t,b_t,h_t,c_t),
\end{equation}
where $s_t$ is NMSE or a monotone score derived from it, $v_t$ is a validity indicator, $b_t$ records whether the candidate improves the current best, $h_t$ stores failure information such as runtime errors or parameter out-of-bounds events, and $c_t$ stores complexity or diagnostic-state information. A-SR uses this feedback for three updates: process-state inference, role utility adaptation, and memory routing.

\paragraph{Role-conditioned prompt context.}
The LLM context can be written as
\begin{equation}
    C_t = \mathrm{Compose}(P,r_t,\mathcal{M}_t,e_{<t},B_t),
\end{equation}
where $P$ is the task specification, $r_t$ is the active role, $\mathcal{M}_t$ is routed memory, $e_{<t}$ is recent evaluator feedback, and $B_t$ is the current best candidate summary. In practice, this context contains the task description, allowed variables, role instruction, current best formula, recent failures, and selected memory snippets.

\section{Coordination Details}

\paragraph{Protocol archetypes.}
The four coordination protocols summarize recurring trajectory regimes observed during search. $\pi_{\mathrm{EGC}}$ performs Exploration-Guided Coordination, $\pi_{\mathrm{RGC}}$ performs Reliability-Guided Coordination, $\pi_{\mathrm{PGC}}$ performs Phase-Guided Coordination, and $\pi_{\mathrm{SGC}}$ performs Stability-Guided Coordination. The selector compares the early reliability and productivity profile to these archetypes and chooses the closest coordination bias. Together, the protocols specify how A-SR allocates roles and routes evidence when the search is productive, reliability-limited, phase-dependent, or stagnant.

\paragraph{Online role-policy adaptation.}
Each role has a utility $U_t(r)$ updated from evaluator-derived rewards. The reward combines validity, best-improvement, invalidity, and parameter-failure signals:
\begin{equation}
    R_t =
    \lambda_v \mathbb{I}_{\mathrm{valid}}
    + \lambda_b \mathbb{I}_{\mathrm{best}}
    - \lambda_i \mathbb{I}_{\mathrm{invalid}}
    - \lambda_p \mathbb{I}_{\mathrm{param}}.
\end{equation}
The selected role utility is updated by exponential smoothing and clipping. A-SR-Static disables this online utility update but keeps protocol selection and routed memory. This makes A-SR-Static a useful variant for isolating the effect of online role-policy adaptation.

\paragraph{State-conditioned memory routing.}
The memory router maps a process state and active role to a context subset:
\begin{equation}
    C_t^{\mathrm{mem}}=\mathrm{RouteMemory}(\mathcal{M}_t,z_t,r_t).
\end{equation}
Invalid-heavy states expose failure memory and validity diagnostics to \emph{Reviewer} and \emph{Simplifier}. Stagnant-valid states expose elite and motif memory to \emph{Generator} and \emph{Analyst}. Complex-valid states expose compression cues to \emph{Simplifier} and validity checks to \emph{Reviewer}. Mixed states use a balanced context chosen by the selected protocol.

\section{Implementation and Distillation Details}

\paragraph{Prompt fields.}
The prompt fields used by A-SR include: task description, allowed variables, current best formula and score, active role, role instruction, recent feedback, elite memory, failure memory, motif memory, validity diagnostics, and role history. These fields are semantic rather than tied to a single string template; prompt surface forms can differ across LLM backbones while preserving the same routed information.

\begin{table}[t]
\centering
\small
\setlength{\tabcolsep}{4pt}
\renewcommand{\arraystretch}{1.08}
\begin{tabular*}{\columnwidth}{@{\extracolsep{\fill}}ll@{}}

\toprule
\textbf{Group} & \textbf{Content} \\
\midrule
Task & Description, variables, units \\
Role & Active role and instruction \\
Search state & Best, feedback, role history \\
Routed memory & Elite, failure, motif, diagnostics \\
Output format & Executable program constraint \\
\bottomrule
\end{tabular*}
\caption{Prompt context groups in A-SR.}
\label{tab:appendix_prompt_groups}
\end{table}

\begin{figure*}[t]
    \centering
    \includegraphics[width=\linewidth]{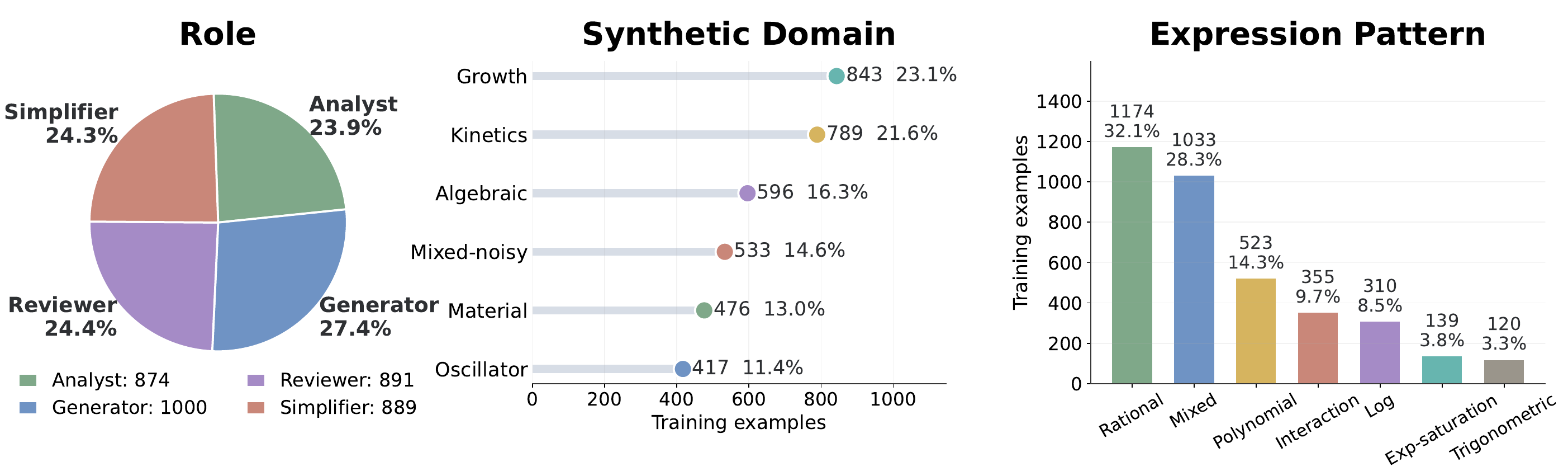}
    \caption{Distribution of the A-SR-LoRA distillation corpus. The figure summarizes only examples used for LoRA training, showing role composition, synthetic-domain coverage, and expression-pattern coverage.}
    \label{fig:sft_data_distribution}
\end{figure*}
\paragraph{Role-specialized instructions.}
The four A-SR roles share the same executable output constraint but differ in the scientific behavior they are asked to perform. The \emph{Generator} proposes new structural hypotheses and explores alternative functional forms. The \emph{Analyst} inspects residual trends, missing terms, and mismatch patterns, then proposes structure changes motivated by those diagnostics. The \emph{Simplifier} attempts to remove redundant terms, reduce unnecessary nesting, and preserve accuracy with a more compact expression. The \emph{Reviewer} focuses on validity, variable usage, numerical stability, and extrapolation-risk proxies. In A-SR and A-SR-Static, these roles are implemented as role-conditioned LLM
calls over a shared backbone; the distinction
comes from role instructions, routed context, and the coordinator's state-dependent
selection. In A-SR-LoRA, the same role interface is further distilled into four
role-specific LoRA modules over the shared Qwen3-4B backbone.

\paragraph{Prompt-template abstraction.}
Prompt templates are organized in a condensed block format rather than only as natural-language descriptions. Each template contains five blocks: task specification, current best candidate, recent evaluator feedback, routed memory, and role-specific instruction. This makes the method reproducible while still allowing minor surface-form changes across LLM backbones. We avoid relying on hidden chain-of-thought outputs; the expected response is an executable candidate equation program or a direct repair of a previous program.

\paragraph{Prompt examples.}
Table~\ref{tab:appendix_prompt_example} shows the canonical prompt layout used by A-SR. The exact surface wording can vary by backbone, but all variants preserve the same information contract: the model receives the task, allowed variables, role, current search state, routed process memory, and executable-output constraint. The response is restricted to a candidate equation program or a direct repair; hidden chain-of-thought is not requested or evaluated.

\paragraph{Role-specific instruction sketches.}
The \emph{Generator} receives broader exploration instructions and is encouraged to introduce alternative functional families when the search is reliable but stagnant. The \emph{Analyst} receives residual, motif, and stability-risk summaries and is asked to propose mechanism-oriented structural edits. The \emph{Simplifier} receives elite and complexity cues and is asked to remove redundant terms without sacrificing validation performance. The \emph{Reviewer} receives failure and validity diagnostics and is asked to repair syntax, variable usage, numerical instability, or extrapolation risks. These role instructions are short by design; most of the role specialization comes from the routed evidence shown to each role.

\paragraph{A-SR-LoRA distillation data.}
\label{app:asr_lora_data}

A-SR-LoRA distills A-SR search behavior into four role-specific LoRA modules
for the shared Qwen3-4B backbone. Each adapter is trained from the subset of
trajectory-distillation examples associated with one A-SR role: Generator,
Analyst, Simplifier, or Reviewer. The corpus is constructed with GPT-5.5 as
the teacher model on a synthetic meta-SR task bank, rather than on
LLM-SRBench validation, test, or OOD labels. Each training example contains a symbolic-regression task specification, active role, role-specific instruction, routed process context, evaluator diagnostics, and an equation-program body proposed under one of the A-SR roles.

We organize the corpus around the four proposal behaviors used by A-SR: Generator, Analyst, Reviewer, and Simplifier. We retain examples that improve the best score, repair a previous failure, reduce complexity without large score degradation, remain valid under the evaluator, or otherwise match the intended role behavior. After deduplication and role balancing, the final training split contains 3,654 examples, with a separate validation split of 400 examples.

Figure~\ref{fig:sft_data_distribution} summarizes the final training split. The role distribution is approximately balanced across the four agent roles, while the synthetic tasks cover growth, reaction kinetics, material constitutive relations, oscillator-like systems, noisy mixed systems, and generic algebraic forms. The resulting equations span diverse expression patterns, including rational, mixed, polynomial, interaction, logarithmic, saturation, and trigonometric structures. This diversity is intended to distill a general proposal prior rather than memorize any target benchmark task. At inference time, the A-SR coordinator selects the active role and dispatches
the corresponding LoRA adapter, while the evaluator, protocol selector, role
policy, and memory router remain unchanged.

\begin{table*}[t]
\centering
\small
\setlength{\tabcolsep}{5pt}
\renewcommand{\arraystretch}{1.08}
\begin{tabular*}{\textwidth}{@{\extracolsep{\fill}}p{0.16\textwidth}p{0.78\textwidth}@{}}
\toprule
\textbf{Method} & \textbf{Hyperparameter Configurations} \\
\midrule
Overall
& LLM temperature $T=0.8$ for default baseline sampling; maximum tokens $=8192$; execution timeout $=30$s per hypothesis; SciPy BFGS constant refinement; maximum 10 free parameters per equation skeleton. \\
LLMDirect
& 5 equation-program hypotheses sampled from the initial prompt; no iterative memory; best valid candidate selected by evaluator score. \\
LLM-SR
& Batch size $b=4$ equation programs per prompt; parallel evaluators $e=4$; islands $m=10$; in-context parent samples $k=4$; experience buffer in prompt context; SciPy BFGS for parameter optimization. \\
LASR
& Iterations $=25$; cycles per iteration $=550$; populations $=10$; population size $=33$; operators: $+, -, \times, \div, \wedge, \exp, \log, \sqrt{\cdot}, \sin, \cos, \tan, \cosh$; LLM weights $=0.005$; remaining configuration follows PySR defaults. \\
SGA
& MSE-driven objective for agentic feedback; differentiable parameter optimization with \texttt{torch.optim.Adam}; PyTorch implementation using \texttt{torch.nn.Module}. \\
DE
& Islands $m=4$; population capacity $=400$ per island; reset interval $=50$; budget $=100$ generations; offspring batch size $=4$; Boltzmann initial temperature $=0.5$; cooling rate $=0.95$; stagnation temperature $=2.0$; parent expressions $k=1$ ($k=2$ for crossover); maximum refinement rounds $=4$; reflective-memory update interval $=12$ generations. \\
\bottomrule
\end{tabular*}
\caption{Implementation details and hyperparameters for LLM-based baselines.}
\label{tab:appendix_baseline_settings}
\end{table*}

\begin{table*}[t]
\centering
\small
\setlength{\tabcolsep}{5pt}
\renewcommand{\arraystretch}{1.08}
\begin{tabular*}{\textwidth}{@{\extracolsep{\fill}}p{0.16\textwidth}p{0.78\textwidth}@{}}
\toprule
\textbf{Domain} & \textbf{Hyperparameter Configurations} \\
\midrule
Search Budget
& Maximum evaluated candidates $B=400$ per task; early profiling budget $B_0=80$; parallel evaluators $e=4$; maximum 10 free parameters per equation skeleton; invalid or non-finite programs rejected before score aggregation. \\
LLM Generation
& Open-source backbones served with vLLM; default sampling temperature $T=0.8$; policy-dependent sampling allowed for protocol-controlled variants; output constrained to executable equation programs; constants fitted externally. \\
Coordination Protocols
& Four protocols: Exploration-Guided, Reliability-Guided, Phase-Guided, and Stability-Guided Coordination; protocol selected after profiling from evaluator-observed reliability and productivity signals. \\
Role Policy
& Four roles: Generator, Analyst, Simplifier, and Reviewer; A-SR-Static disables online role-utility updates; A-SR updates role utilities after $B_0$ using evaluator-derived rewards. \\
Process Memory
& Elite memory stores top valid candidates; failure memory stores invalid programs and error types; motif memory stores recurring structures; diagnostics store validity, complexity, and parameter-fitting signals. \\
A-SR-LoRA
& Trajectory distillation applied to Qwen3-4B-Instruct; role-specific LoRA modules trained from filtered A-SR traces; inference uses the same A-SR evaluator and coordinator. \\
\bottomrule
\end{tabular*}
\caption{Implementation details and hyperparameters in A-SR.}
\label{tab:appendix_asr_settings}
\end{table*}

\paragraph{Evaluation protocol.}
Continuous equation parameters are optimized with BFGS-style quasi-Newton optimization \citep{fletcher2013practical}, consistent with compared LLM-guided SR baselines. LLM inference for open-source backbones is served with vLLM \citep{kwon2023vllm}. Unless otherwise stated, train and validation metrics are averaged for main LLM-SRBench results, while OOD metrics are reported separately in the main text and supplementary analysis.

\section{Experimental Configuration and Reproducibility}

\paragraph{Backbone models.}
The main benchmark experiments use open-source instruction-tuned backbones, including Llama3.1-8B-Instruct and Qwen3-4B-Instruct-2507. The real-world scientific tasks use GPT-3.5 to align with the reported LLM-SR setting. We separate results by backbone because the absolute quality of LLM-generated equation programs can vary substantially across models, and because A-SR is intended as a coordination framework rather than a single-backbone symbolic regressor.

\paragraph{Search budget.}
All A-SR variants are evaluated under a fixed candidate-evaluation budget for each task. The budget counts evaluated formulas rather than raw LLM tokens. This convention matches prior LLM-guided SR work, where the main cost is the number of candidate programs sent through parsing, parameter fitting, and numerical evaluation. When comparing against methods with reported results, we only make direct claims under matched or clearly stated budget assumptions.

\paragraph{A-SR variants.}
A-SR-Static keeps hierarchical protocol selection and role-aware process memory but disables online role-policy adaptation. A-SR additionally updates role utilities using evaluator-derived rewards during search. A-SR-LoRA uses a trajectory-distilled open-source backbone inside the same A-SR coordination loop. These variants are reported separately because they answer different questions: whether hierarchical coordination helps without online adaptation, whether online role-policy updates add further benefit, and whether search trajectories can improve an open-source agent prior.
\begin{table*}[t]
\centering
\small
\setlength{\tabcolsep}{3pt}
\renewcommand{\arraystretch}{1.05}
\begin{tabular*}{\textwidth}{@{\extracolsep{\fill}}lcccccccc@{}}
\toprule
\textbf{Method} &
\multicolumn{2}{c}{\textbf{Physics}} &
\multicolumn{2}{c}{\textbf{Material}} &
\multicolumn{2}{c}{\textbf{Chemistry}} &
\multicolumn{2}{c}{\textbf{Biology}} \\
\cmidrule(lr){2-3}
\cmidrule(lr){4-5}
\cmidrule(lr){6-7}
\cmidrule(lr){8-9}
& \textbf{OOD NMSE} & \textbf{Acc@0.01}
& \textbf{OOD NMSE} & \textbf{Acc@0.01}
& \textbf{OOD NMSE} & \textbf{Acc@0.01}
& \textbf{OOD NMSE} & \textbf{Acc@0.01} \\
\midrule
\multicolumn{9}{c}{\textbf{Qwen3-4B-Instruct-2507}} \\
\midrule
LLMDirect
& 3.43e4 & 6.82
& 1.94e-1 & 72.00
& 5.98e6 & 0.00
& 1.18e2 & 4.17 \\
LLM-SR
& 8.17e4 & 9.09
& 2.75e-1 & 72.00
& 6.57e2 & 11.11
& 2.24e1 & 12.50 \\
LASR
& 1.08e4 & 9.09
& 3.25e-1 & 40.00
& 4.49e2 & 2.78
& 9.66e2 & 4.17 \\
SGA
& 9.92e5 & 6.82
& 2.91e-1 & 40.00
& 1.55e3 & 2.78
& 1.83e5 & 4.17 \\
DE
& 1.97e3 & 15.91
& 4.85e-2 & 80.00
& 1.09e1 & 11.11
& 1.14e1 & 12.50 \\
A-SR-Static
& 2.340e1 & 22.73
& 8.148e-4 & 96.00
& 4.035e0 & 30.56
& 5.725e1 & 25.00 \\
\textbf{A-SR}
& 3.171e2 & 18.18
& 3.641e-3 & 96.00
& 8.490e-1 & 19.44
& 7.408e0 & 16.67 \\
\textbf{A-SR-LoRA}
& 4.319e3 & 25.00
& 6.438e2 & 88.00
& 7.500e-1 & 27.78
& 1.948e0 & 25.00 \\
\midrule
\multicolumn{9}{c}{\textbf{Llama3.1-8B-Instruct}} \\
\midrule
A-SR-Static
& 4.347e2 & 25.00
& 1.391e-3 & 96.00
& 1.566e0 & 22.22
& 2.022e-1 & 37.50 \\
\textbf{A-SR}
& 6.552e3 & 25.00
& 6.897e-4 & 100.00
& 2.059e0 & 25.00
& 1.077e0 & 45.83 \\
\bottomrule
\end{tabular*}
\caption{Out-of-distribution evaluation on LSR-Synth. Metrics include OOD NMSE ($\downarrow$) and OOD Acc@0.01 ($\uparrow$, \%).}
\label{tab:ood_appendix}
\end{table*}

\begin{table*}[t]
\centering
\small
\setlength{\tabcolsep}{3pt}
\renewcommand{\arraystretch}{1.05}
\begin{tabular*}{\textwidth}{@{\extracolsep{\fill}}lcccccc@{}}
\toprule
\textbf{Setting} & \textbf{Train NMSE} & \textbf{Train Acc} & \textbf{Val NMSE} & \textbf{Val Acc} & \textbf{OOD NMSE} & \textbf{OOD Acc} \\
\midrule
fixed role rotation & 7.37e-2 & 24.44 & 7.62e-2 & 24.44 & 3.08e0 & 24.44 \\
w/o routed memory & 9.02e-2 & 26.67 & 9.12e-2 & 26.67 & 4.87e-1 & 24.44 \\
w/o adaptive protocol selection & \textbf{5.37e-2} & 26.67 & \textbf{5.61e-2} & 26.67 & 3.51e0 & \underline{33.33} \\
A-SR-Static & \underline{6.21e-2} & \underline{28.89} & \underline{5.88e-2} & \underline{28.89} & \textbf{1.03e-1} & \underline{33.33} \\
\textbf{A-SR online} & 7.00e-2 & \textbf{33.33} & 7.62e-2 & \textbf{31.11} & \underline{3.67e-1} & \textbf{35.56} \\
\bottomrule
\end{tabular*}
\caption{Full ablation results on the mixed 45-task subset of LLM-SRBench using Llama3.1-8B-Instruct. Metrics include NMSE ($\downarrow$) and Acc@0.01 ($\uparrow$, \%). Bold and underlined values indicate the best and second-best performance, respectively.}
\label{tab:appendix_mixed45_ablation}
\end{table*}

\paragraph{Baseline settings.}
We summarize implementation details and hyperparameters for LLM-guided symbolic-regression baselines following the configuration style used by DE. Direct generation uses a best-of-$N$ prompt without iterative process memory. LLM-SR-style proposal search uses iterative equation-program generation with external constant fitting and an experience buffer. LASR and SGA are included when compatible LLM-SRBench settings are available. DE is treated as the closest agentic baseline and is included for matched LLM-SRBench settings. Table~\ref{tab:appendix_baseline_settings} summarizes baseline configurations, and Table~\ref{tab:appendix_asr_settings} summarizes the corresponding A-SR configuration.

For the four real-world LLM-SR scientific-discovery tasks, DE is not included in the main comparison because it does not provide a complete matched result table for those tasks. This choice avoids mixing directly comparable reported baselines with incomplete or non-reproducible settings.

\paragraph{Metrics.}
We report normalized mean squared error (NMSE) and Acc@0.01. NMSE measures numerical fit, while Acc@0.01 summarizes the fraction of tasks solved below a fixed error threshold. We include both because average NMSE can be dominated by a small number of catastrophic failures, whereas Acc@0.01 highlights whether a method reliably finds sufficiently accurate formulas across tasks. For real-world tasks, OOD NMSE is treated as a separate generalization metric rather than averaged into ID performance.
\begin{figure*}[t]
\centering
\includegraphics[width=\textwidth]{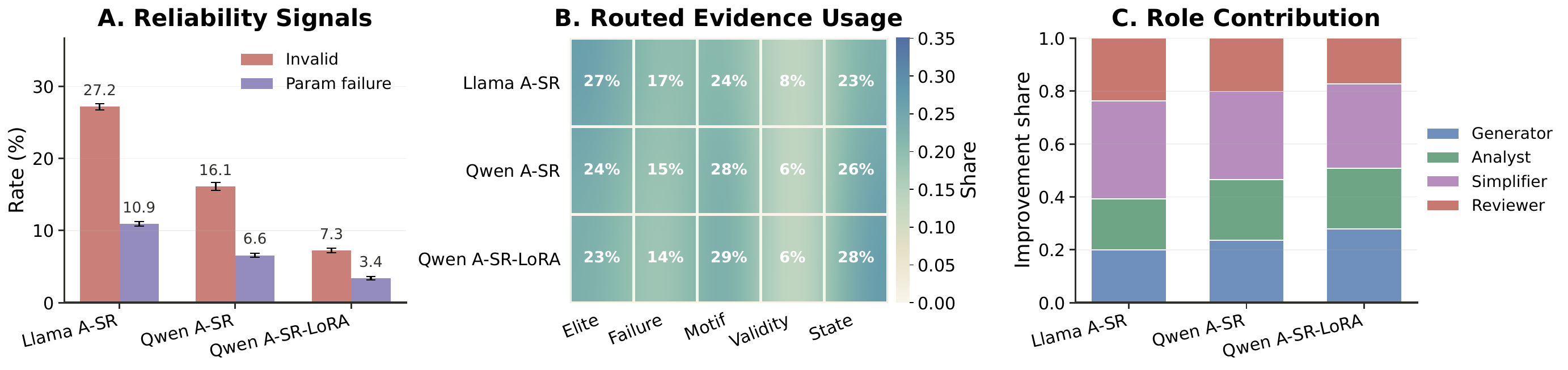}
\caption{Aggregate process diagnostics from A-SR runs. Panel A reports invalid-program and parameter-failure rates. Panel B shows routed evidence usage: Elite, Failure, and Motif are persistent memory blocks, while Validity and State are diagnostic/context blocks derived from evaluator feedback and inferred process state. Panel C reports the fraction of best-improvement events attributed to each role.}
\label{fig:appendix_compact_diagnostics}
\end{figure*}
\begin{figure*}[t]
\centering
\includegraphics[width=\textwidth]{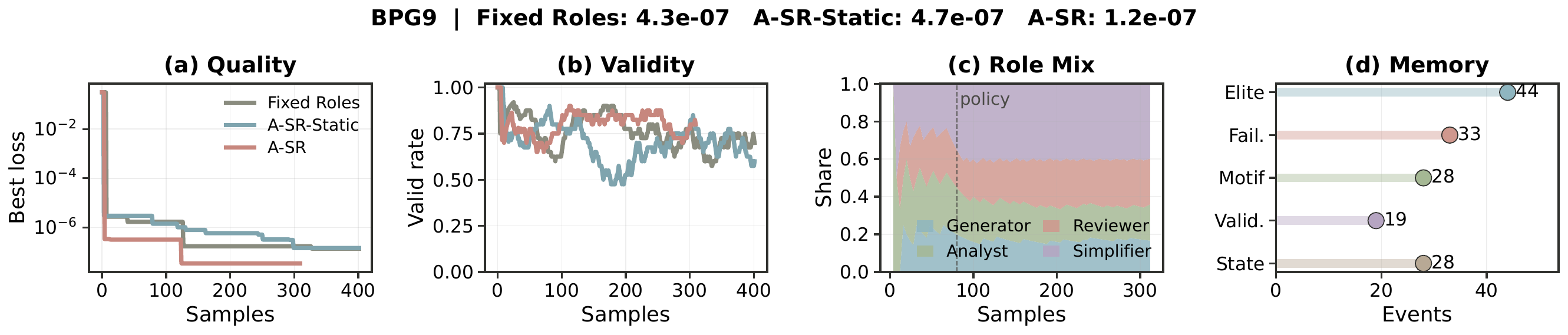}
\caption{Additional search dynamics on the Biology task BPG9. The panels follow the same layout as Figure~\ref{fig:best_score_trajectories}: best-so-far loss, rolling valid-candidate rate, role distribution, and memory-routing events.}
\label{fig:appendix_search_dynamics_biology}
\end{figure*}


\section{Additional Experimental Results}

\paragraph{OOD per-domain results.}
Table~\ref{tab:ood_appendix} reports the full out-of-distribution evaluation for LSR-Synth groups with held-out extrapolation splits. The main LLM-SRBench table reports train/validation aggregate performance, so this appendix table reports only OOD NMSE and OOD Acc@0.01. We separate OOD results from the main aggregate table to avoid conflating interpolation fit with extrapolation robustness, which is especially important for scientific symbolic regression.

\paragraph{Noise and backbone robustness.}
Symbolic regression is sensitive to observation noise because flexible expressions can explain noise through unnecessary nonlinear terms, and because continuous parameter fitting can become ill-conditioned when the candidate skeleton is over-parameterized. A-SR is not a noise-specific symbolic-regression method, but several design choices reduce common noise-induced failures: the \emph{Simplifier} receives complexity cues, the \emph{Reviewer} receives numerical-stability diagnostics, and the coordinator separates valid-but-stagnant states from invalid-heavy states. In the current experiments, we evaluate noisy behavior through the real-world LLM-SR tasks and through OOD splits rather than a full synthetic noise sweep.

A-SR is intended as a coordination framework rather than a single-model symbolic regressor. The benchmark experiments use Llama3.1-8B-Instruct and Qwen3-4B-Instruct-2507, while the real-world scientific-discovery setting uses GPT-3.5 to match the original LLM-SR protocol. These backbones differ in absolute proposal quality, but the evaluator, process-state inference, role routing, memory routing, and online role-utility update are model-agnostic. The comparisons among A-SR-Static, A-SR online, and A-SR-LoRA therefore isolate coordination behavior from raw backbone strength.

\paragraph{Ablation details.}
Table~\ref{tab:appendix_mixed45_ablation} provides the full NMSE and Acc@0.01 values for the mixed 45-task ablation summarized in the main text. The main text emphasizes Acc@0.01 because mean NMSE can be dominated by a small number of catastrophic symbolic failures, while the appendix reports both error and solved-task reliability. These results show that fixed role rotation is not sufficient by itself, and that removing routed memory or adaptive protocol selection reduces reliability.

Table~\ref{tab:appendix_bio_ablation} reports a focused Biology-subset ablation using the same component-removal settings. This subset is useful because online role-policy adaptation gives a clear improvement over the static coordinator on both validation and OOD accuracy. The Biology result therefore complements the mixed-subset ablation by showing that the full online coordinator is especially helpful when search states change across valid, stagnant, and extrapolation-sensitive regimes.

\begin{table*}[t]
\centering
\small
\setlength{\tabcolsep}{3pt}
\renewcommand{\arraystretch}{1.05}
\begin{tabular*}{\textwidth}{@{\extracolsep{\fill}}lcccccc@{}}
\toprule
\textbf{Setting} & \textbf{Train NMSE} & \textbf{Train Acc} & \textbf{Val NMSE} & \textbf{Val Acc} & \textbf{OOD NMSE} & \textbf{OOD Acc} \\
\midrule
fixed role rotation
& 1.297e-2 & 8.33
& 1.187e-2 & 8.33
& 6.862e1 & 8.33 \\
w/o role-aware memory routing
& 7.480e-3 & 20.83
& 7.153e-3 & 20.83
& 1.671e1 & 20.83 \\
w/o adaptive protocol selection
& 2.451e-2 & 25.00
& 2.461e-2 & 25.00
& 8.386e1 & 25.00 \\
A-SR-Static / w/o online
& \underline{8.241e-4} & \underline{41.67}
& \underline{9.081e-4} & \underline{41.67}
& \textbf{2.022e-1} & \underline{37.50} \\
\textbf{A-SR online}
& \textbf{5.396e-6} & \textbf{50.00}
& \textbf{5.358e-6} & \textbf{50.00}
& \underline{1.077e0} & \textbf{45.83} \\
\bottomrule
\end{tabular*}
\caption{Ablation results on the Biology subset of LLM-SRBench using Llama3.1-8B-Instruct. Metrics include NMSE ($\downarrow$) and Acc@0.01 ($\uparrow$, \%). Bold and underlined values indicate the best and second-best performance, respectively.}
\label{tab:appendix_bio_ablation}
\end{table*}

\begin{table*}[t]
\centering
\small
\setlength{\tabcolsep}{3pt}
\renewcommand{\arraystretch}{1.08}
\begin{tabular}{p{0.17\textwidth}p{0.16\textwidth}p{0.15\textwidth}p{0.20\textwidth}p{0.24\textwidth}}
\toprule
\textbf{Stage} & \textbf{Roles} & \textbf{State} & \textbf{Sketch} & \textbf{Effect} \\
\midrule
Early profiling
& Multi-role
& Mixed
& Simple valid forms
& Estimate reliability/productivity \\
Exploration
& Gen./Anal.
& Stagnant-valid
& Nonlinear variants
& Route elite/motif memory \\
Repair
& Rev./Simp.
& Invalid-heavy
& Validity fixes
& Route failure diagnostics \\
Compression
& Simp./Rev.
& Complex-valid
& Compact variants
& Route compression cues \\
Refinement
& Gen./Anal.
& Productive
& Best-motif variants
& Continue controlled search \\
\bottomrule
\end{tabular}
\caption{Schematic discovery stages for role selection and memory routing in A-SR.}
\label{tab:appendix_discovery_trace}
\end{table*}


\section{Formula-Level Case Study Details}
\label{app:formula_case_details}

Table~\ref{tab:case_formula_comparison} reports representative cases where A-SR recovers expressions that are algebraically equivalent to the reference after scalar constant fitting. The fitted outputs shown in the table substitute the optimized constants into the model expressions, making the recovered signs and multiplicative factors explicit. These examples show that A-SR can recover the reference dependency structure rather than only achieving low aggregate error.

Tables~\ref{tab:case_study_logged_i107} and~\ref{tab:case_crk16} provide formula-level case studies. We report the benchmark reference expression, the model output, routed-memory evidence, coordination traces, and evaluation metrics.

\section{Failure Mode Analysis}

\paragraph{Invalid programs.}
Invalid programs include syntax errors, undefined variables, unsupported operations, shape mismatches, and numerical exceptions. A-SR handles this failure mode by routing failure memory and validity diagnostics to \emph{Reviewer} and \emph{Simplifier}, increasing the probability that subsequent proposals repair executable structure before pursuing additional exploration.

\paragraph{Parameter out-of-bounds and numerical instability.}
Some generated equation skeletons are syntactically valid but lead to unstable constant fitting or parameter values outside the allowed range. These cases are treated separately from generic invalidity because they often indicate that the symbolic structure is too ill-conditioned or over-parameterized. A-SR exposes these traces to roles that can simplify the expression or replace unstable substructures.

\paragraph{Stagnation under valid formulas.}
A common failure mode in SR is that the search repeatedly produces valid formulas with similar scores but no structural improvement. A-SR treats this as a different state from invalid-heavy failure. In stagnant-valid regimes, the coordinator routes elite formulas and recurring motifs to \emph{Generator} and \emph{Analyst}, encouraging structural exploration grounded in what has already worked.

\paragraph{Over-complex formulas.}
Accurate but overly complex formulas may fit training data while generalizing poorly. A-SR does not use compactness as the sole objective, but it exposes complexity cues to \emph{Simplifier} and \emph{Reviewer} so that redundant terms can be removed when they do not provide clear predictive benefit. This makes compactness part of the coordination signal rather than a separate post-processing step.

\paragraph{Extrapolation risk.}
A candidate can fit the training and validation splits while remaining structurally fragile under distribution shift. A-SR does not access held-out OOD labels during search. Instead, it exposes evaluator-observable proxies such as numerical instability, parameter-fitting failures, excessive complexity, and unstable equivalent forms to \emph{Reviewer} and \emph{Analyst}. These signals do not directly optimize OOD error, but they encourage candidates that are executable, stable, and less dependent on brittle over-parameterization.

\section{Additional Search Dynamics and Process Diagnostics}

Figure~\ref{fig:appendix_search_dynamics_biology} extends the main-text CRK12 analysis with a representative Biology trace. The figure uses complete 400-sample traces. This example illustrates how A-SR changes role choices after the profiling phase, routes multiple memory types during search, and improves the best-so-far trajectory in a domain where the full online coordinator also improves aggregate validation and OOD accuracy.

\paragraph{Schematic discovery stages.}
Table~\ref{tab:appendix_discovery_trace} summarizes the main coordination regimes used by A-SR. The table is schematic rather than a single-task transcript: it shows how evaluator-observed states are associated with role choices and routed memory views.

\paragraph{Aggregate process diagnostics.}
Figure~\ref{fig:appendix_compact_diagnostics} summarizes process-level signals from the A-SR runs. Panel A reports reliability signals used by the coordinator, including invalid-program and parameter-failure rates. Panel B reports routed evidence usage. Here, Elite, Failure, and Motif correspond to persistent process-memory blocks, while Validity and State denote diagnostic/context blocks derived from evaluator feedback and inferred process state. Explicit validity evidence is routed selectively, mainly under invalid or parameter-failure states; in productive searches, the controller more frequently uses elite, motif, and state evidence. Panel C reports the fraction of best-improvement events attributed to each role, showing that improvement is distributed across roles rather than dominated by a single prompt variant.

\paragraph{Task-level search dynamics.}
The task-level trace in Figure~\ref{fig:appendix_search_dynamics_biology} illustrates how these aggregate process patterns appear within an individual search. The biology trace shows steady improvement with active memory routing and non-collapsed role usage across the search budget. We treat this figure as qualitative process evidence; aggregate comparisons are reported in Tables~\ref{tab:main_llmsrbench} and~\ref{tab:appendix_bio_ablation}.

\section{Ethics Statement}
This work studies symbolic regression methods on public or synthetic benchmark datasets. The experiments do not involve human subjects, private user data, personally identifiable information, or sensitive decision-making applications. The main ethical considerations are reproducibility and faithful reporting of experimental results.

\section{LLM Usage Statement}
We used LLM-based writing assistance for grammar polishing and clarity editing of the manuscript. An LLM was also used to construct the synthetic trajectory-distillation corpus for A-SR-LoRA. It was not used to inspect or label benchmark validation/test/OOD targets.

\begin{table*}[t]
\centering
\small
\setlength{\tabcolsep}{5pt}
\renewcommand{\arraystretch}{1.16}
\begin{tabularx}{\textwidth}{p{0.18\textwidth}X}
\toprule
\textbf{Component} & \textbf{Case Study Content -- Example 1} \\
\midrule

Problem &
\textbf{Task:} \texttt{I.10.7\_1\_0}. Predict velocity $v$ from relativistic mass $m$, rest mass $m_0$, and speed of light $c$. \\

Reference expression &
\[
v^\star = -c\sqrt{1-\frac{m_0^2}{m^2}} .
\] \\[-1.0em]

Routed memory excerpt &
\begin{minipage}[t]{\linewidth}
\logtext{[productive focus] Preserve useful motifs from elite candidates while varying one structural component.}

\vspace{2pt}
\logtext{[motif memory] Useful operator motifs: *++(10), *+++-+/+pow+sqrt(2), *+++/+pow+sin+sqrt(2), *+-+/+pow+sqrt(2).}

\vspace{2pt}
\logtext{[failure memory] Recent failed candidate patterns: execution failed or returned invalid score; execution failed or returned invalid score; execution failed or returned invalid score.}
\end{minipage} \\

Coordination evidence &
\begin{minipage}[t]{\linewidth}
\logtext{Selected policy: Stability-Guided Coordination.}

\vspace{2pt}
\logtext{States observed in routed memory: mixed, productive, complex\_but\_valid, invalid\_high.}

\vspace{2pt}
\logtext{Role trace: Analyst 20; Simplifier 36; Reviewer 28; Generator 16.}

\vspace{2pt}
\logtext{Search dynamics: initial best score = -4.163; final best score = -2.330e-13; best-score improvements = 12.}
\end{minipage} \\

Model output &
\begin{minipage}[t]{\linewidth}
\logtext{return params[0] * c * (1 - (m\_0 / m)**2)**0.5}
\end{minipage} \\

Fitted expression &
\[
 p_0\approx -1
\] \\[-1.0em]

Evaluation &
Train NMSE $=5.60{\times}10^{-14}$, Val NMSE $=2.87{\times}10^{-13}$; Acc@0.01 is satisfied on both train and validation. \\

\bottomrule
\end{tabularx}
\caption{Case study for \texttt{I.10.7\_1\_0} with routed-memory excerpts, role trace, model output, and search statistics.}
\label{tab:case_study_logged_i107}
\end{table*}

\begin{table*}[t]
\centering
\small
\setlength{\tabcolsep}{5pt}
\renewcommand{\arraystretch}{1.16}
\begin{tabularx}{\textwidth}{p{0.18\textwidth}X}
\toprule
\textbf{Component} & \textbf{Case Study Content -- Example 2} \\
\midrule

Problem &
\textbf{Task:} \texttt{chem\_react\_CRK16}. Predict the reaction rate $\mathrm{d}A/\mathrm{d}t$ from time $t$ and concentration $A$. \\

Reference expression &
The benchmark metadata specifies the ground-truth skeleton
\[
\frac{\mathrm{d}A}{\mathrm{d}t}=k_f A^{1/3}-kA .
\]
\[
k_f\approx0.970929,\qquad k\approx0.245812 .
\] \\[-1.0em]

Routed memory excerpt &
\begin{minipage}[t]{\linewidth}
\logtext{[productive focus] Preserve useful motifs from elite candidates while varying one structural component.}

\vspace{2pt}
\logtext{[elite memory] -3.433e-13: def equation(t, A, params): ... Improved version of equation\_v1 with additional nonlinear terms ...}

\vspace{2pt}
\logtext{[motif memory] Useful operator motifs: *+++-+/+exp+pow+sin(10), *+++-+/+exp+sin(10), *+++-+/+cos+exp+pow+sin(8), *+++-+exp+pow+sin(6).}
\end{minipage} \\

Coordination evidence &
\begin{minipage}[t]{\linewidth}
\logtext{Selected policy: Phase-Guided Coordination.}

\vspace{2pt}
\logtext{State: productive; memory blocks: state\_focus:productive, elite\_memory, motif\_memory.}

\vspace{2pt}
\logtext{Role trace at final logged stage: Generator 62; Analyst 20; Reviewer 13; Simplifier 5.}

\vspace{2pt}
\logtext{Online utility updates include valid=true, reward=0.2 and param\_oob=true, reward=-0.8.}
\end{minipage} \\

Model output &
\begin{minipage}[t]{\linewidth}
\logtext{rate\_term = params[0] * A ** params[1] * (1 + np.sin(4 * t)) ** params[2]}

\vspace{2pt}
\logtext{decay\_term = -params[3] * A}

\vspace{2pt}
\logtext{activation\_term = params[4] * A}

\vspace{2pt}
\logtext{return rate\_term + decay\_term + activation\_term}
\end{minipage} \\

Fitted expression &
\[
p_0=0.970929,\quad
p_1=0.333328,\quad
p_2=-1.89{\times}10^{-8},\quad
p_3=1.114426,\quad
p_4=0.868621 ,
\]
\[
\hat f(t,A)\approx0.970929A^{1/3}-0.245805A .
\] \\[-1.0em]

Evaluation &
Train NMSE $=7.95{\times}10^{-12}$, Val NMSE $=8.97{\times}10^{-12}$, OOD NMSE $=2.77{\times}10^{-4}$; Acc@0.01 is satisfied on train, validation, OOD. \\

\bottomrule
\end{tabularx}
\caption{Case study for \texttt{chem\_react\_CRK16} with routed-memory excerpts, role trace, model output, and fitted effective expression.}
\label{tab:case_crk16}
\end{table*}

\clearpage
\end{document}